\documentclass[10pt,twocolumn,letterpaper]{article}

\usepackage{etex} %
\usepackage{cvpr}
\usepackage{times}
\usepackage{epsfig}
\usepackage{graphicx}
\usepackage{amsmath}
\usepackage{amssymb}

\usepackage{alex}
\usepackage{booktabs} %
\usepackage{subcaption} %
\usepackage{algorithm} %
\usepackage{algpseudocode} %
\usepackage{rotating}
\usepackage{enumitem}
\usepackage[font={small}]{caption}

\usepackage{pgfplots}
\usepackage{tango}
\usepackage{tikz}
\usetikzlibrary{calc,positioning,arrows}
\usetikzlibrary{automata,positioning}
\usetikzlibrary{fit} %
\usetikzlibrary{backgrounds} %
\usetikzlibrary{positioning}
\usetikzlibrary{shadows}
\usetikzlibrary{bayesnet}
\usepgflibrary{shapes}
\tikzstyle{observed}=[circle, thick, minimum size=0.9cm, draw=black!100, fill=black!20]
\tikzstyle{void}=[circle, thick, minimum size=0.9cm, draw=black!0, fill=black!0]
\tikzstyle{lstm}=[rectangle, minimum size=0.2cm, draw=black!100]
\tikzstyle{shadeplate}=[rectangle, thick, inner sep=0.4cm, draw=black!100]
\tikzstyle{table}=[circle,fill=blue!20,draw=black!100,inner sep=1pt, minimum size=30pt]
\tikzstyle{client}=[rectangle,fill=blue!20,draw=black!100,inner sep=1pt, minimum size=12pt]

\usepgfplotslibrary{colorbrewer}
\definecolor{Dark2-8-1}{RGB}{27,158,119}
\definecolor{Dark2-8-A}{RGB}{27,158,119}
\definecolor{Dark2-8-2}{RGB}{217,95,2}
\definecolor{Dark2-8-B}{RGB}{217,95,2}
\definecolor{Dark2-8-3}{RGB}{117,112,179}
\definecolor{Dark2-8-C}{RGB}{117,112,179}
\definecolor{Dark2-8-4}{RGB}{231,41,138}
\definecolor{Dark2-8-D}{RGB}{231,41,138}
\definecolor{Dark2-8-5}{RGB}{102,166,30}
\definecolor{Dark2-8-E}{RGB}{102,166,30}
\definecolor{Dark2-8-6}{RGB}{230,171,2}
\definecolor{Dark2-8-F}{RGB}{230,171,2}
\definecolor{Dark2-8-7}{RGB}{166,118,29}
\definecolor{Dark2-8-G}{RGB}{166,118,29}
\definecolor{Dark2-8-8}{RGB}{102,102,102}
\definecolor{Dark2-8-H}{RGB}{102,102,102}

\graphicspath{{../images/}}

\usepackage[breaklinks=true,bookmarks=false]{hyperref}
\hypersetup{
     colorlinks   = false,
     citecolor    = green
}
\hypersetup{linkcolor=red}
\pgfplotsset{compat = 1.3}

\cvprfinalcopy %

\def\cvprPaperID{3021} %

\newcommand{\KG}{\textcolor{black}} %
\newcommand{\KGtwo}{\textcolor{black}} %
\newcommand{\cc}{\textcolor{black}} %
\newcommand{\ccc}{\textcolor{black}} %
\newcommand{\CHECK}{\textcolor{black}} %

\newcommand{\myparagraph}[1]{\textbf{#1}\ }

\ifcvprfinal\fi
\begin{document}

\title{Creating Capsule Wardrobes from Fashion Images}

\author{Wei-Lin Hsiao\\
UT-Austin\\
{\tt\small kimhsiao@cs.utexas.edu}
\and
Kristen Grauman\\
UT-Austin\\
{\tt\small grauman@cs.utexas.edu}
}

\maketitle
\begin{abstract}%
  We propose to automatically create \emph{capsule wardrobes}.  Given an inventory of candidate garments and accessories, the algorithm must assemble a minimal set of items that provides maximal mix-and-match outfits.  We pose the task as a subset selection problem.  To permit efficient subset selection over the space of all outfit combinations, we develop submodular objective functions capturing the key ingredients of visual compatibility, versatility, and user-specific preference.  Since adding garments to a capsule only expands its possible outfits, we devise an iterative approach to allow near-optimal submodular function maximization.  Finally, we present an unsupervised approach to learn visual compatibility from ``in the wild" full body outfit photos; the compatibility metric  translates well to cleaner catalog photos and improves over existing methods.  Our results on thousands of pieces from popular fashion websites show that automatic capsule creation
  has potential to mimic skilled fashionistas in assembling flexible wardrobes, while being significantly more scalable.
\end{abstract}%

\section{Introduction}
\vspace{-1mm}

The fashion domain is a magnet for computer vision.
New vision problems are emerging in step with the fashion industry's rapid
evolution towards an online, social, and personalized business.
Style models~\cite{hipster,128floats,snavely-street-style,weilin-iccv2017,style2vec}, trend forecasting~\cite{ziad-iccv2017}, interactive search~\cite{augmented-memory,whittle-search}, and recommendation~\cite{mcauley-dyadic,davis-mm2015,magic-closet} all require visual understanding with rich detail and subtlety.  Research in this area is poised to have great influence on
what people buy, how they shop, and how the fashion industry analyzes its enterprise.

\begin{figure}[t]\vspace{-1mm}
\centering
\includegraphics[width=\linewidth]{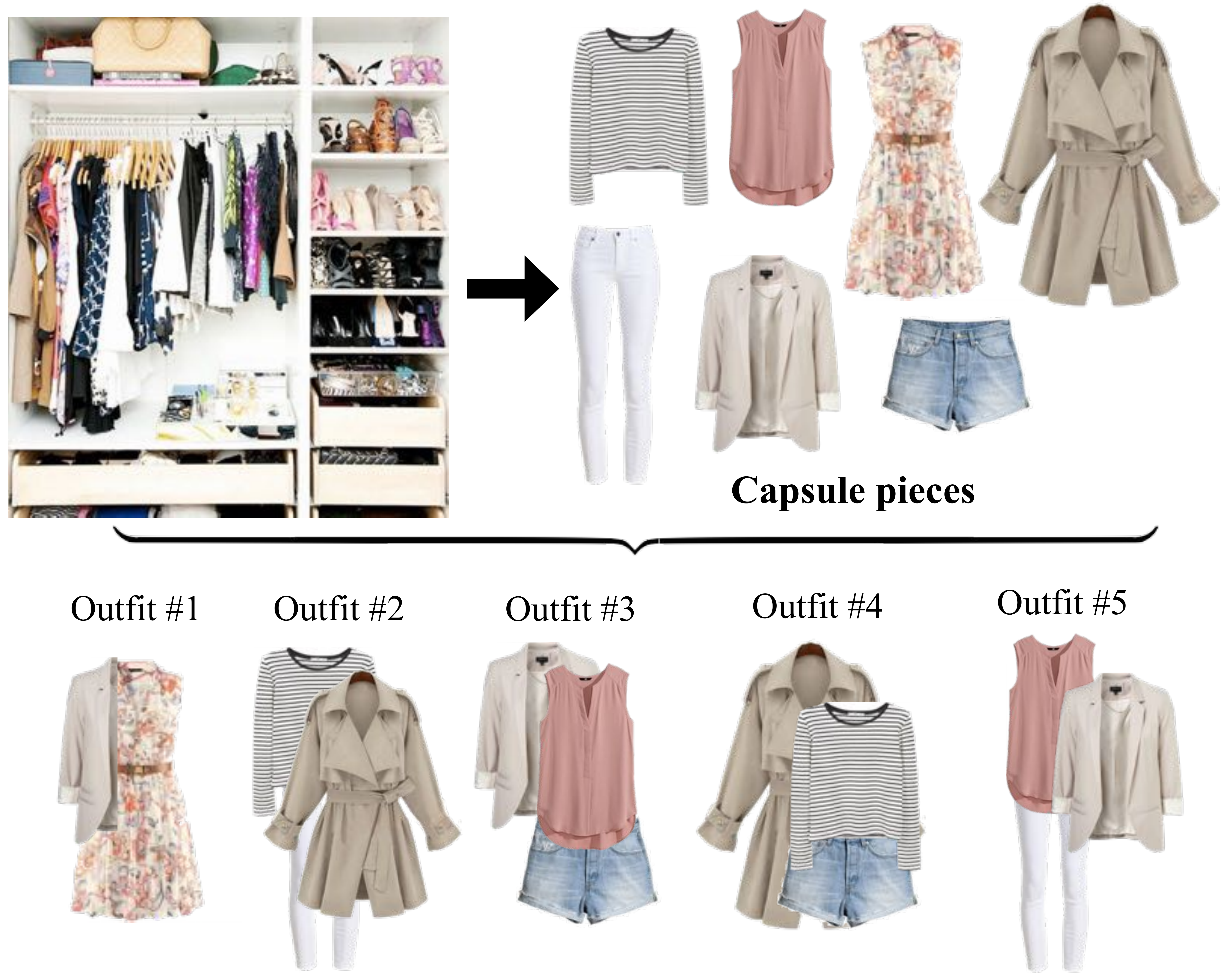}\vspace*{-0.1in}
\caption{A \emph{capsule wardrobe} is a minimal set of garments that can mix and match to compose many visually compatible outfits.}%
 \label{fig:concept}
\end{figure}

A \emph{capsule wardrobe} is a set of garments that can be assembled into many visually compatible outfits (see \figref{concept}).  Capsules are currently the purview of fashion experts, magazine editors, and bloggers.  A capsule creator manually analyzes an inventory to puzzle together a relatively small set of clothing items and accessories that mix and match well.  A curated capsule can help consumers get the best value for their dollars (``do more with less"), help vendors propose appealing wardrobes from their catalogs, and help a subscription service (e.g., StitchFix, LeTote) ship a targeted box that amplifies a customer's wardrobe.

We propose to automate capsule wardrobe generation.  There are two key technical challenges.  First, capsules hinge on having an accurate model of \emph{visual compatibility}.  Whereas visual \emph{similarity} asks ``what looks like this?", and is fairly well understood~\cite{fu2012efficient,street-to-shop2012,runway-realway,lin2015rapid},  \emph{compatibility} instead asks ``what complements this?"  It requires capturing how multiple visual items interact, often according to subtle visual properties.  Existing compatibility methods~\cite{mcauley-dyadic,bilstm,mcauley-style,mcauley-compatibility,magic-closet,neurostyle}
assume supervision via labels or co-purchase data, which limits scope and precision, as we will discuss in \secref{approach}.
Furthermore, they largely cater only to pairwise compatibility.  %
The second challenge is that capsule generation is a complex combinatorial problem.  Of all possible garments, we seek the subset that maximizes versatility and compatibility, and, critically, the addition of any one garment introduces \emph{multiple} new outfit combinations. %

We introduce an approach to automatically create capsule wardrobes that addresses both of these issues.
We first cast capsule creation as a subset selection problem.  We define an objective function characterizing a capsule based on its pieces' mutual compatibility, the resulting outfits' versatility, and (optionally) its faithfulness to a user's preferred style.  Then, we develop an efficient algorithm that maps a large inventory of candidate garments into the best capsule of the desired size.  We design objectives that are submodular for the addition of new \emph{outfits}, ensuring the ``diminishing returns'' property that facilitates near-optimal set selection~\cite{submodular,leskovec-cost}.  Then, since each \emph{garment} added to a capsule \emph{expands} the possible outfits, we further develop an iterative approach that exploits outfit submodularity to alternate between fixing and selecting each layer of clothing.  

As a second main contribution, we introduce an unsupervised approach to learn visual compatibility from full-body images ``in the wild".  We learn a generative model for outfit compositions from unlabeled images that can score $k$-way compatibility.  %
Because it is built on predicted attributes, our model can translate compatibility learned from the ``in the wild" photos to cleaner catalog photos of individual items, where users need most guidance on mixing and matching.

We evaluate our approach on thousands of garments from Polyvore, popular social commerce websites for fashion.  
We compare our algorithm's capsule creations to those manually defined by fashionistas, as well as subjective user studies.  Furthermore, we show our underlying compatibility model offers advantages over some state of the art methods.  Finally, we demonstrate the practical value of our algorithm, which in seconds finds near-optimal capsules for problem scales that are otherwise intractable.

\section{Related Work}
\myparagraph{Attributes for fashion}
Attributes offer a natural representation for clothing, since they can describe relevant patterns (\emph{checked}, \emph{paisley}), colors (\emph{rose}, \emph{teal}), fit (\emph{loose}), and cut (\emph{V-neck}, \emph{flowing})~\cite{gallagher-eccv2012,bossard2012apparel,di2013style,mixmatch2015,ddan,kim2016visual,deepfashion}.
Topic models on attributes are indicative of styles~\cite{iwata,vaccaro,weilin-iccv2017}.
\KG{Inspired by~\cite{weilin-iccv2017},
we employ topic models.  However, whereas~\cite{weilin-iccv2017} seeks a style-coherent image embedding, we use correlated topic models to score novel combinations of garments for their compatibility.}
Domain adaptation~\cite{ddan,darn} and multi-task curriculum learning~\cite{mtct} are valuable to overcome the gap between street and shop photos.  We devise a simple curriculum learning approach to train attributes effectively in our setting.  None of the above methods explore visual compatibility or capsule wardrobes.

\myparagraph{Style and fashionability}
Beyond recognition tasks, fashion also demands answering: How do we represent \emph{style}?  What makes an outfit \emph{fashionable}?
The style of an outfit is typically learned in a supervised manner.
Leveraging style-labeled data like HipsterWars~\cite{hipster} or DeepFashion~\cite{deepfashion}, classifiers built on body keypoints~\cite{hipster}, weak meta-data~\cite{128floats}, or contextual embeddings~\cite{style2vec} show promise.
Fashionability refers specifically to a style's popularity.  It can also be learned from supervised data, e.g., 
online data for user ``likes"~\cite{fashion-compose,fashionability}.  
Unsupervised style discovery methods instead mine unlabeled photos to detect common themes in people's outfits, with topic models~\cite{weilin-iccv2017}, non-negative matrix factorization~\cite{ziad-iccv2017}, or clustering~\cite{snavely-street-style}.  We also leverage unlabeled images to discover ``what people wear"; however, our goal is to infer visual compatibility for unseen garments, rather than trend analysis~\cite{ziad-iccv2017,snavely-street-style} or image retrieval~\cite{weilin-iccv2017} on a fixed corpus.

\myparagraph{Compatibility and recommendation}
Substantial prior work explores ways to link images containing the \emph{same} or very similar garment~\cite{fu2012efficient,street-to-shop2012,runway-realway,getting-the-look2013,lin2015rapid}.
In contrast, compatibility requires judging how well-coordinated or \emph{complementary} a given set of garments is.
Compatibility can be posed as a metric learning problem%
~\cite{mcauley-dyadic,mcauley-style,mcauley-compatibility},
addressable with Siamese embeddings~\cite{mcauley-dyadic} or link prediction~\cite{mcauley-style}.  
Text data can aid compatibility~\cite{fashion-compose,neurostyle,bilstm}.
As an alternative to metric learning, a recurrent neural network models 
outfit composition as a sequential process that adds one garment at a time, implicitly learning compatibility via the transition function~\cite{bilstm}.  Compatibility has applications in recommendation~\cite{magic-closet,davis-mm2015}, but prior work recommends a garment at a time, as opposed to constructing a wardrobe.

To our knowledge, all prior work requires labeled data to learn compatibility, whether from human annotators curating matches~\cite{bilstm,davis-mm2015}, co-purchase data~\cite{mcauley-dyadic,mcauley-style,mcauley-compatibility}, or implicit crowd labels~\cite{fashion-compose}.  %
In contrast, we propose an unsupervised approach, which has the advantages of scalability, privacy, and continually refreshable models as fashion evolves, and also avoids awkwardly generating ``negative'' training pairs (see \secref{approach}).  
Most importantly, our work is the first to develop an algorithm for generating capsule wardrobes.  Capsules require going beyond pairwise compatibility to represent $k$-way interactions and versatility, and they present a challenging combinatorial problem.

\myparagraph{Subset selection}
We pose capsule wardrobe generation as a subset selection problem.  %
Probabilistic determinantal point processes (DPP) can identify the subset of items that maximize individual item ``quality" while also maximizing total ``diversity" of the set~\cite{dpp}, and have been applied for document and video summarization~\cite{dpp,dpp-video}.  Alternatively, submodular function maximization exploits ``diminishing returns" to select an optimal subset subject to a budget~\cite{submodular}.  For submodular objectives, an efficient greedy selection criterion is near optimal~\cite{submodular}, e.g., as exploited for sensor placement~\cite{krause-sensor-placement} and outbreak detection~\cite{leskovec-cost}.  We show how to adapt such solutions to permit accurate and efficient selection for capsule wardrobes; furthermore, we develop an iterative EM-like algorithm to enable non-submodular objectives for mix-and-match outfits.
\section{Approach}
\vspace{-1.7mm}
\label{sec:approach}
We first formally define the capsule wardrobe problem and introduce our approach (\secref{capsule_wardrobe}). Then in \secref{CTM} we present our unsupervised approach to learn \emph{compatibility} and \emph{personalized styles}, two key ingredients in capsule wardrobes. Finally, in \secref{attr_recognition}, we overview our training procedure for cross-domain attribute recognition.

\subsection{Subset selection for capsule wardrobes}
\label{sec:capsule_wardrobe}
\vspace{-0.5mm}
A capsule wardrobe is a minimal set of garments that combine in versatile ways to create many compatible outfits (see \figref{concept}). We cast capsule creation as the problem of selecting a subset from a large set of candidates that maximizes quality (compatibility) and diversity (versatility).
\vspace*{-0.1in}
\subsubsection{Problem formulation and objective}
\vspace*{-0.05in}

We formulate the subset selection problem as follows.
Let $i=0,\dots,(m-1)$ index the $m$ layers of clothing (e.g.,~outerwear, upper body, lower body, hosiery).
Let $A_i = \cbr{s_{i}^{0}, s_{i}^{1}, ..., s_{i}^{N_i-1}}$ denote the set of candidate \cc{garments/pieces} in layer $i$, where $s_{i}^{j}$, $j=0,\dots,(N_i-1)$ is the $j$-th piece in layer $i$, and $N_i$ is the number of candidate pieces for that layer. For example, the candidates could be the inventory of a given catalog.
If an outfit is composed of \textit{one and only one} piece from each layer, the candidate pieces in total could generate a set $\mathcal{Y}$ of $\prod_i {N_i}$  possible outfits.

\begin{figure} %
    \vspace{-3mm}
    \begin{center}
         \includegraphics[width=\linewidth]{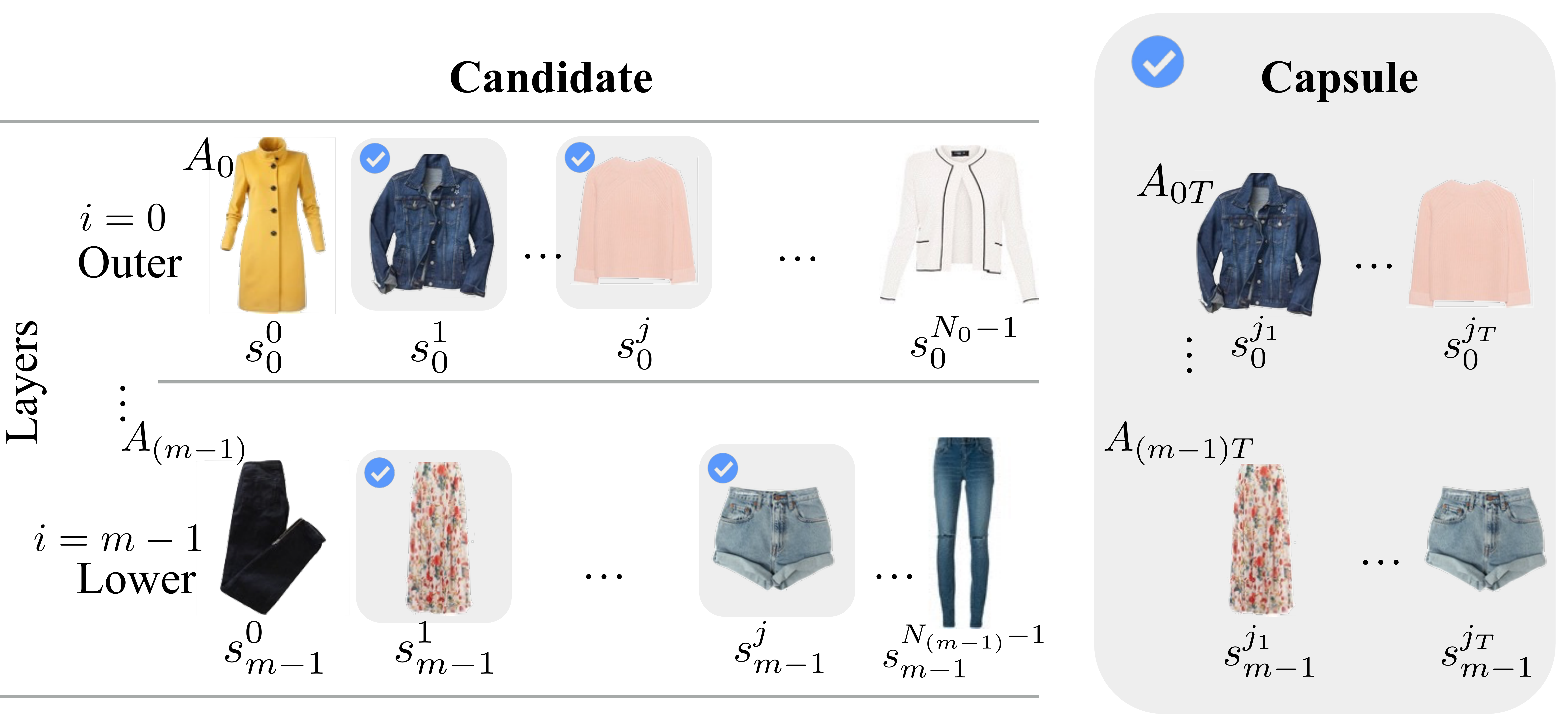}
    \end{center}\vspace*{-0.2in}
    \caption{\small{Selecting a subset of pieces from the candidates to form a capsule wardrobe. Left shows candidates from all layers. The selected pieces are checked and compose the subset on the right.}}
    \label{fig:problem_form}\vspace{-4mm}
 \end{figure}

\myparagraph{Objective}
To form a capsule wardrobe, we must select only $T$ pieces, $A_{iT} = \cbr{s_{i}^{j_1}, ..., s_{i}^{j_{T}}} \subseteq A_i$ from each layer $i$.
The set of outfits $\mathbf{y}$ generated by these pieces consists of $A_{0T}\times~A_{1T}\times\ldots\times A_{(m-1)T}$. Our goal is to select the pieces $A_{iT}^*, \forall i$ such that their composed set of outfits $\mathbf{y^*}$ is maximally \emph{compatible} and \emph{versatile}.  \figref{problem_form} visualizes this problem.

To this end, we define our objective as:
\begin{equation}
\begin{aligned}
\mathbf{y^*} = \argmax_{\mathbf{y} \subseteq \mathcal{Y}} C(\mathbf{y})+ V(\mathbf{y}), \\
s.t.~\mathbf{y} = A_{0T}\times~A_{1T}\times\ldots\times A_{(m-1)T}
\end{aligned}
\end{equation}
where $C(\mathbf{y})$ and $V(\mathbf{y})$ denote the compatibility and versatility scores, respectively.  

A na\"\i ve approach to find the optimal solution $\mathbf{y^*}$ requires computation on $T^m$ outfits in a subset, multiplying by $\binom{N}{T}^m$ to search through all possible subsets.
Since our candidate pool may consist of all merchandise in a shopping site, $N$ may be on the order of hundreds or thousands, so optimal solutions become intractable. 
Fortunately, our key insight is that as wardrobes expand, subsequent outfits add diminishing amounts of new styles/looks.  This permits a \textit{submodular} objective that allows us to obtain a near-optimal solution efficiently.
In particular, greedily growing a set for subset selection is near-optimal if the objective function is submodular; the greedy algorithm is guaranteed to reach a solution achieving at least a constant fraction $1-\frac{1}{e}$, or about 63\%, of the optimal score~\cite{submodular}.
\vspace*{-0.05in}
\begin{definition}{Submodularity.}
A set function $F$ is submodular if, $\forall D \subseteq B \subseteq V$, $\forall s \in V \setminus B$, $F(D\cup\cbr{s})-F(D) \geq F(B\cup\cbr{s})-F(B)$.
\end{definition}
\vspace*{-0.05in}
Submodularity satisfies \textit{diminishing returns}.  Since it is closed under nonnegative linear combinations, if we design $C(\mathbf{y})$ and $V(\mathbf{y})$ to be submodular, our final objective will be submodular as well, as we show next.

We stress that \emph{items} in a subset $\mathbf{y}$ are \emph{outfits}---not garments/pieces.
The capsule is the Cartesian product of all garments selected per layer.
Therefore, when greedily growing a set at time step $t$, an incremental addition of one garment $s_{i}^{j_t}$ from layer $i$ entails adding $s_{i}^{j_t}\times\prod_{i' \neq i} A_{i'(t-1)}$ new outfits.  %
While ultimately the algorithm must add garments to the capsule,
for the sake of optimization, it needs to reason about the set \emph{in terms of those garments' combinatorial outfits}.
We address this challenge below.

\myparagraph{Compatibility}
Suppose we have an algorithm that returns the compatibility estimate \KG{$c(o_j)$} of outfit $o_j$ (to be defined in Sec.~\ref{sec:CTM}).  We define a set's compatibility score as:
\vspace*{-0.07in}
\begin{equation}
C(\mathbf{y}) := \Sigma_{o_j \in \mathbf{y}} \KG{c(o_j)},
\label{eq:compatibility}
\vspace*{-0.01in}
\end{equation}
the sum of compatibility scores for all its outfits.  $C(\mathbf{y})$ is modular, %
a special case of submodularity, since an additional outfit $o_j$ to any set $\mathbf{y}$ will increase $C(\mathbf{y})$ by the same amount $c(o_j)$.

\myparagraph{Versatility} A good capsule wardrobe should offer a variety of looks, or styles, for different uses and occasions.  %
We formalize versatility as a \textit{coverage} function over all styles:
\vspace*{-0.05in}
\begin{equation}
V(\mathbf{y}) := \Sigma_{i=1}^{K} v_\mathbf{y}\KG{(z_i)}
\label{eq:coverage}
\vspace*{-0.05in}
\end{equation}
where $v_\mathbf{y}(z_i)$ measures the degree to which outfits in $\mathbf{y}$ cover the $i$-th desired style $z_i$, and $K$ is the total number of distinct styles.  Sec.~\ref{sec:CTM} will define our model for styles $z_i$.
To satisfy the diminishing returns property, we define $v_\mathbf{y}(i)$ probabilistically:
\vspace*{-0.1in}
\begin{equation}
v_\mathbf{y}\KG{(z_i)} := 1 - \prod_{o_j \in \mathbf{y}} (1-P(z_i|o_j)),\label{eq:coverageprob}%
\end{equation}
where $P(z_i | o_j)$ denotes the probability of a style $z_i$ given an outfit $o_j$.  We define a generative model for $P(z_i|o_j)$ below.  The idea is that each outfit ``tries" to cover a style with probability $P(z_i | o_j)$, and the style is covered by a capsule if at least one of the outfits in it successfully covers that style.  Thus, as $\mathbf{y}$ expands, subsequent outfits add diminishing amounts of style coverage.  \KG{A probabilistic expression for coverage is also used in~\cite{blogosphere} for blog posts.}

We have thus far defined versatility in terms of uniform coverage over \emph{all} $K$ styles. However, each user has his/her own preferences, and a universally versatile capsule may contain pieces that do not meet one's taste. Thus, as a personalized variant of our approach, we adjust each style's proportion in the coverage function by a user's style preference.  This extends our capsules with \emph{personalized versatility}:
\begin{equation}
\KG{V^\prime}(\mathbf{y}) := \Sigma_{i=1}^{K} w_i v_\mathbf{y}(z_i),
\label{eq:personal_coverage}
\end{equation}
where $w_i$ denotes a personalized preference for each style $i$.  \KG{Sec.~\ref{sec:CTM} explains how the personalization weights are discovered from user data.}

\vspace*{-0.15in}
\subsubsection{Optimization}
\label{sec:optimization}
\vspace*{-0.04in}

A key challenge of subset selection for capsule wardrobes is that our subsets are on \emph{outfits}, but we must form the subset by selecting \emph{garments}. With each \cc{garment} addition, the subset of outfits $\mathbf{y}$ grows superlinearly, since every new \cc{garment} can combine with all previous \cc{garments} to form new outfits.  
\cc{Submodularity requires each addition to diminish a set function's gain, but
  adding more garments yields \emph{more} outfits, so the gain actually increases.}
Thus, \KGtwo{while our objective is submodular for adding outfits, it is not submodular for adding individual garments}.
However, we can make the following claim:
\vspace*{-0.07in}
\begin{claim} When fixing all other layers (i.e., upper, lower, outer) and selecting a subset of pieces one layer at a time, the probabilistic versatility coverage function in \eqnref{coverage} is submodular, and the compatibility function in \eqnref{compatibility} is modular.  See Supplementary File for proof.
\label{claim_submodular}
\end{claim}
\vspace*{-0.07in}

Thus, given a single layer, our objective function is submodular \KGtwo{for garments}. By fixing all selected pieces in other layers, any additional \cc{garment} will be combined with the same set of \cc{garments} and form the same amount of new outfits.  Thus subsets in a given layer no longer grow superlinearly.
So the guarantee of a greedy solution on that layer being near-optimal~\cite{submodular} still holds.

To exploit this, we develop an EM-like iterative approach to approximate a greedy solution over all layers: we iteratively fix the subsets selected in other layers, and focus the current selection in a single layer.  After sufficient iterations, our subsets converge to a fixed set. Algorithm \ref{em_greedy_alg} gives the complete steps.

\begin{center}
\begin{algorithm}[t]
    \footnotesize
    \caption{Proposed iterative greedy algorithm for submodular maximization, \KG{where $\mathbf{obj}(\mathbf{y}) := C(\mathbf{y}) + V(\mathbf{y})$.}}
    \label{em_greedy_alg}
    \begin{algorithmic}[1]
        \State $A_{iT} := \varnothing, \forall i$
        \State $\Delta_{obj} := \varepsilon + 1$ \Comment{$\varepsilon$ is the tolerance degree for convergence}
        \State $\mathbf{obj}^{m-1}_{prev} := 0$
        \While {$\Delta_{obj}^{m-1}\geq\varepsilon$}
            \For {each layer $i=0,1,...(m-1)$}
                \State $A_{iT} = A_{i0} := \varnothing$ \Comment{Reset selected pieces in layer $i$}
                \State $\mathbf{obj}^{i}_{cur} := 0$
                \For {each time step $t=1,2,...T$}
                    \State $\mathbf{y_{t-1}} = A_{i(t-1)}\times\prod_{i' \neq i} A_{i'T}$
                    \State $s_{i}^{j_t} := \argmax_{s \in A_i \setminus A_{i(t-1)}} \delta_s$ \Comment{Max increment}
                    \State where $\delta_s = \mathbf{obj}(\mathbf{y_{t-1}}\uplus s) - \mathbf{obj}(\mathbf{y_{t-1}})$
                    \State $A_{it} := s_{i}^{j_t}\cup A_{i(t-1)}$ \Comment{Update layer $i$}
                    \State $\mathbf{obj}^{i}_{cur} := \mathbf{obj}^{i}_{cur} + \delta_{s_{i}^{j_t}}$
                \EndFor
            \EndFor
            \State $\Delta_{obj}^{m-1} := \mathbf{obj}^{m-1}_{cur}-\mathbf{obj}^{m-1}_{prev}$
            \State $\mathbf{obj}^{m-1}_{prev} := \mathbf{obj}^{m-1}_{cur}$
        \EndWhile
        \Procedure{Incremental Addition }{$\mathbf{y_t} := \mathbf{y_{t-1}}\uplus s$}
        \State $\mathbf{y_{t}^+} := s, s \in A_i \setminus A_{i(t-1)}$
        \For {$j \in \cbr{1,\dots,m}, j\neq i$}
            \If {$A_{jT}\neq\varnothing$}
                \State $\mathbf{y_{t}^+} := \mathbf{y_{t}^+}\times A_{jT}$
            \EndIf
        \EndFor
        \State $\mathbf{y_{t}} := \mathbf{y_{t-1}} \cup \mathbf{y_{t}^+}$
        \EndProcedure
    \end{algorithmic}
\end{algorithm}\vspace{-4mm}
\end{center}\vspace{-4mm}

Our algorithm is quite efficient.  \KGtwo{Whereas a na\"\i ve search would take more than 1B hours for our data with $N=150$, our algorithm returns an approximate capsule in only 200 seconds.}  Most computation is devoted to computing the objective function, which requires topic model inference (see below).  A na\"\i ve greedy approach \KGtwo{on \cc{garments}} would require $O(NT^4)$ time for $m=4$ layers, while our iterative approach requires $O(NT^3)$ time per iteration (details in Supp.) For our datasets, it requires just $5$ iterations.  
\subsection{Style topic models for compatibility}
\label{sec:CTM}

Having defined the capsule selection objective and optimization, we now present our approach to model versatility (via $P(z_i|o_j)$) and compatibility $c(o_j)$ simultaneously.

\begin{figure}
   \begin{center}
    \begin{subfigure}{.12\textwidth}
        \includegraphics[width=\linewidth]{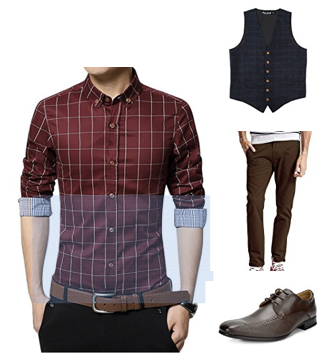}
    \end{subfigure}\hspace*{0.15in}\hfill
    \begin{subfigure}{.14\textwidth}
        \includegraphics[width=\linewidth]{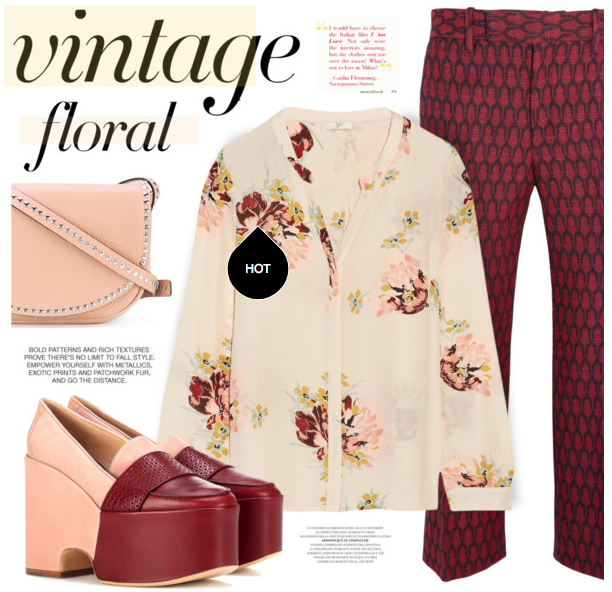}
    \end{subfigure}\hspace*{0.15in}\hfill
    \begin{subfigure}{.09\textwidth}
        \includegraphics[width=\linewidth]{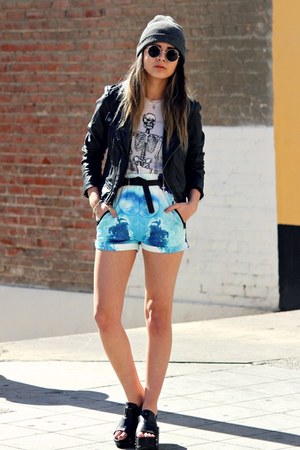}
    \end{subfigure}
    \end{center}
   \vspace*{-0.25in}
   \caption{L to R: Amazon co-purchase example; Polyvore user-curated set; Chictopia full-body outfit (our training source).}
   \label{fig:amazon}
   \vspace*{-4mm}
\end{figure}

Prior work on compatibility takes a supervised approach.  Given ground truth compatible items (either by manual labels like curated sets of product images on Polyvore~\cite{fashion-compose,neurostyle,bilstm} or by using Amazon co-purchase data as a proxy~\cite{mcauley-style,mcauley-dyadic,mcauley-compatibility}), a (usually discriminative) compatibility metric is trained.  See ~\figref{amazon}.
However, the supervised strategy has weaknesses.  First,
items purchased at the same time can be a weak proxy for visual compatibility.  %
Second, user-created sets often focus on the visualization of the collage, usually contain fewer than two main (non-accessory) pieces, and lack layers like hosiery.  Third, sources like Amazon and Polyvore are limited to brands selected by vendors, a fraction of the wide variety of clothing people wear in real life.
Fourth, obtaining the \emph{negative} non-compatible examples required by supervised discriminative methods is problematic.
Previous work~\cite{mcauley-style,mcauley-dyadic,mcauley-compatibility,fashion-compose,neurostyle} generates negative examples by randomly swapping items in positive pairs, but there is no guarantee that the random combinations are true negatives.
Not observing a pair of items together does not necessarily mean they do not go well together.

To address these issues, we propose a generative compatibility model that is learned from unlabeled images of people wearing outfits ``in the wild" (\figref{amazon}, right).
We explore a topic model---namely Correlated Topic Models (CTM)~\cite{blei-ctm}
---from text analysis. CTM is a Bayesian multinomial mixture model that supposes a small number of $K$ latent \emph{topics} account for the distribution of observed words in any given document. It uses the following generative process for a corpus $\mathcal{D}$ consisting of $M$ documents each of length $L_{i}$:  
\begin{enumerate*}
\item Choose $\boldsymbol{\eta_{i}}\,\sim \,\mathcal{N}(\boldsymbol{\mu},\boldsymbol{\Sigma})$, where $i\in \{1,\dots ,M\}$ and $\boldsymbol{\mu},\boldsymbol{\Sigma}$ a $K$-dimensional mean and covariance matrix$. \\{\theta_{ik}} = \frac{e^{{\eta_{ik}}}}{\Sigma_{k=1}^{K} e^{{\eta_{ik}}}}$ maps $\boldsymbol{\eta_i}$ to a simplex.
\item Choose $\boldsymbol{\varphi _{k}}\,\sim \,\mathrm {Dir} (\beta )$, where $k\in \{1,\dots ,K\}$ and $\mathrm {Dir} (\beta )$ is the Dirichlet distribution with parameter $\beta$.
\item For each word indexed by $(i,j)$,\\ where $j\in \{1,\dots ,L_{i}\}$, and $i\in \{1,\dots ,M\}$.
\begin{enumerate*}
\item Choose a topic $z_{i,j}\,\sim \,\mathrm {Multinomial} (\boldsymbol{\theta _{i}})$.
\item Choose a word $x_{i,j}\,\sim \,\mathrm {Multinomial} (\boldsymbol{\varphi _{z_{i,j}}})$.
\end{enumerate*}
\end{enumerate*}
Only the word occurrences are observed. %
Following~\cite{weilin-iccv2017}, we map textual topic models to visual ones: a ``document" is an outfit, a ``word" is an inferred visual attribute (e.g., \emph{floral}, \emph{chiffon}), and a ``topic" is a style.   The model discovers the compositions of visual cues (i.e.,~attributes) that characterize styles. \cc{A topic might capture plaid blue blouses, or tight leather skirts.}
Prior work~\cite{weilin-iccv2017} models styles with a Dirichlet prior, which treats topics within an image as independent.  For compatibility, we find CTM's logistic normal prior above~\cite{blei-ctm} beneficial to  account for style correlations (e.g., a formal blazer is more likely to be combined with a skirt than sporty leggings.)

CTM estimates the latent variables by maximizing the posterior distribution given a corpus $\mathcal{D}$:
\vspace{-2mm}
\begin{equation}
p(\boldsymbol{\theta},\boldsymbol{z}|\mathcal{D},\boldsymbol{\mu},\boldsymbol{\Sigma},\beta) = \prod_{i=1}^{M}p(\boldsymbol{\theta_i}|\boldsymbol{\mu},\boldsymbol{\Sigma})\prod_{j=1}^{\KG{L_i}}p(z_{ij}|\boldsymbol\theta_i) p(x_{ij}|z_{ij},\beta).
\label{eq:lda_posterior}
\end{equation}
First we find the latent variables that fit assembled outfits on full-body images.  
\cc{Next, given an arbitrary combination of catalog pieces, we predict their attributes, and take the union of attributes on all pieces to form an outfit $o_j$. Finally we infer its compatibility by the likelihood:}
\begin{equation}
c(o_j) := p(o_j | \boldsymbol{\mu},\boldsymbol{\Sigma},\beta).
\end{equation}
Combinations similar to previously assembled outfits $\mathcal{D}$ will have higher probability.  For this generative model, no negative examples need to be contrived.  %
The training pool should be outfits like those we want the model to emulate; we use full-body photos posted to a fashion website.

Given a database of unlabeled outfit images, we predict their attributes.  Then we apply CTM to obtain a set of styles, where each style $k$ is an attribute distribution $\boldsymbol{\varphi_k}$. Given an outfit $o_j$, CTM infers its style composition $\boldsymbol{\theta_{o_j}}=[\theta_{{o_j}1},\dots,\theta_{{o_j}K}]$.  Then we have:
\begin{equation}
P(z_i | o_j) := P(z_i | \boldsymbol{\theta_{o_j}}) = \theta_{{o_j}i},
\end{equation}
which is used to compute our versatility coverage in Eq~\eqref{eq:coverageprob}.

We personalize the styles emphasized for a user in Eq~\eqref{eq:personal_coverage} as follows.  Given a collection of outfits $\{o_{j_1},\dots,o_{j_U}\}$ owned by a user $p$, \KG{e.g., as shown in that user's purchase history catalog photos or his/her album on a fashion website}, we learn his/her style preference $\boldsymbol{\theta_p^{(user)}}$ by aggregating all outfits $\boldsymbol{\theta^{(user)}_{p}} = \frac{1}{U}\Sigma_j \boldsymbol{\theta_{o_j}}$. Hence, a user's personalized weight $w_i$ for each style $i$ is $\theta^{(user)}_{pi}$.

Unlike previous work that uses supervision from human created matches or co-purchase information, our model is fully unsupervised. While we train attribute models on a disjoint pool of attribute labeled images, our topic model runs on ``inferred'' attributes, and annotators do not touch the images from which we learn compatibility.

\begin{figure*}[t] %
   \vspace{-7mm}
   \begin{center}
        \includegraphics[width=\linewidth]{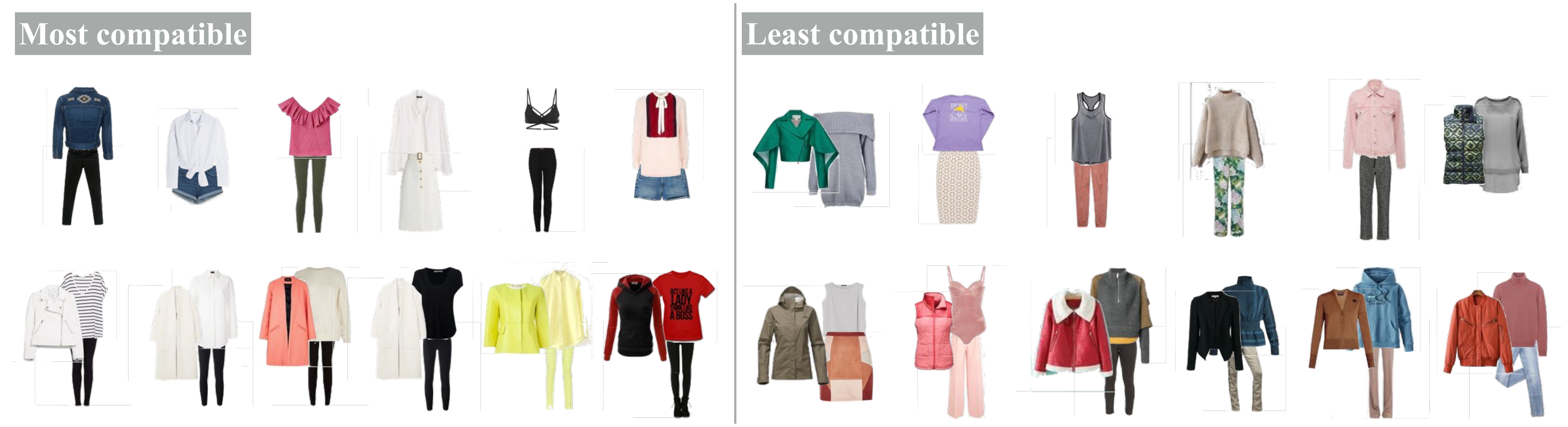}
   \end{center}\vspace*{-0.25in}
   \caption{\small{Qualitative examples of most (left) and least (right) compatible outfits as scored by our model. \KG{Here we show both outfits with $2$ pieces (top) and $3$ pieces (bottom).}}}
   \label{fig:qual_compatibility}\vspace*{-0.15in}
\end{figure*}

\subsection{Cross-domain attribute recognition}
\label{sec:attr_recognition} 

Finally, we describe our approach to infer cross-domain attributes on both catalog and outfit images.

\myparagraph{Vocabulary and data collection}
Since fine-grained attributes are valuable for style, we build our vocabulary on the 195 attributes enumerated in~\cite{weilin-iccv2017}. Their dataset has attributes labeled only on \emph{outfit} images, so we collect \emph{catalog} images labeled with the same attribute vocabulary. To gather images, we use keyword search with the attribute name on Google, then manually prune those where the attribute is absent. This yields 100 to 300 positive training images per attribute, in total \CHECK{12K} images. (Details in Supp.)

\myparagraph{Curriculum learning from shop to street}
Catalog images usually have a clear background, and are free of tricky lighting conditions and deformations from body pose. As a result, attributes are more readily recognizable in catalog images than full-body outfit images.  We propose a two stage curriculum learning approach for cross domain attribute recognition. In the first stage, we finetune a deep neural network, ResNet-50~\cite{resnet} pretrained on ImageNet~\cite{imagenet}, for \emph{catalog attribute} recognition. In the second stage, we first detect and crop images into upper and lower body instances, and then we finetune the network from the first stage for \emph{outfit attribute} recognition.  Evaluating on the validation split from the 19K image dataset~\cite{weilin-iccv2017}, we see a significant $15\%$ mAP improvement, especially on challenging attributes such as \emph{material} and \emph{neckline}, which are subtle or occupy a small region on outfit images.

\section{Experiments}\label{sec:results}

We first evaluate our compatibility estimation in isolation (\secref{exp_compatibility}). Then, we evaluate our algorithm's capsule wardrobes, for both quality and efficiency %
(\secref{exp_capsule}).

\subsection{Compatibility}
\label{sec:exp_compatibility}
\myparagraph{Dataset}
Previous work \cite{fashion-compose,neurostyle,bilstm} studying compatibility each collects a dataset from \url{polyvore.com}.  Polyvore is a platform where fashion-conscious users create sets of clothing pieces that go well with each other.  While these compatible sets are valuable supervision, as discussed above, collecting ``incompatible'' outfits is problematic.

While our method uses no ``incompatible'' examples for training, to facilitate evaluation, we devise a more reliable mechanism to avoid false negatives.  We collect $3,759$ Polyvore outfits, composed of $7,478$ pieces, each with meta-labels such as \emph{season (winter, spring, summer, fall)}, \emph{occasion (work, vacation)}, and \emph{function (date, hike)}.  \tabref{dataset_stats} summarizes the dataset breakdown.  We exploit the meta-labels to generate incompatible outfits.  For each compatible outfit, we generate an incompatible one by randomly swapping one piece to another piece in the same layer %
from an exclusive meta-label.  %
For example, each \emph{winter (work)} outfit will swap a piece with a \emph{summer (vacation)} outfit. We use outfits that have at least $2$ pieces from different layers as positives, and for each positive outfit, we generate $5$ negatives. In total, \KG{our test set} has $2,574$ positives and $12,870$ negatives.  \cc{In short, by swapping with the guidance of meta-label, the negatives are more likely to be true negatives.}  
\KG{See Supp. for examples.}

\begin{table} %
   \centering
   \footnotesize
   \setlength\tabcolsep{2pt}
   \begin{tabular}{@{}lcccc|cc|cc|c@{}}                   
      & fall & winter & spring & summer & vacation & work & date & hike & \bf total \\
      \midrule
      total sets & 307 & 302 & 307 & 308 & 731 & 505 & 791 & 508 & 3759 \\
      $>1$ itm. & 242 & 275 & 227 & 206 & 421 & 454 & 514 & 413 & 2752 \\
      $>2$ itm. & 101 & 130 & 71 & 60 & 66 & 177 & 96 & 146 & 847
   \end{tabular}
   \vspace*{-0.1in}
   \caption{Breakdown of our Polyvore dataset.}  %
   \label{tab:dataset_stats}
   \vspace{-6mm}
\end{table}

\myparagraph{Baselines}
We compare with two recent methods: i) \textsc{Monomer}~\cite{mcauley-compatibility}, an embedding trained using Amazon product co-purchase info as a proxy label for compatibility.  Since Monomer predicts only pairwise scores between pieces, we average all pairwise scores to get an outfit's total compatibility, following~\cite{bilstm}. ii) \textsc{BiLSTM}~\cite{bilstm}, a sequential model trained on user-created sets from Polyvore to predict the held-out piece of a layer given pieces from other layers.  The probability of the whole sequence is its compatibility metric. For both baselines, we use the authors' provided code and their same training data sources.

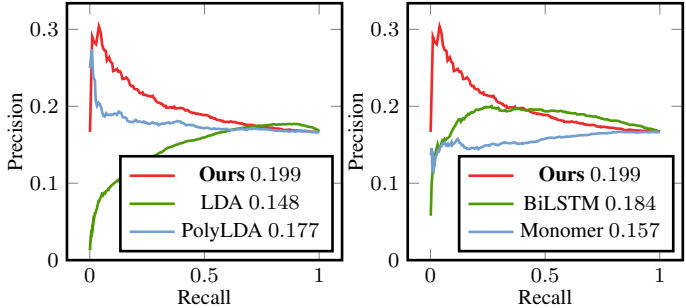
\begin{figure}
\footnotesize
\centering
\hspace*{-0.3in}
\begin{subfigure}[b]{0.3\textwidth} %
\pgfplotsset{every axis/.append style={
    font=\footnotesize,,
    line width=1pt,
    tick style={ultra thin}}}
    \begin{tikzpicture}
        \begin{axis}[
            xlabel=Recall,
            ylabel=Precision,
            ymin = 0,
            y label style={yshift=-3pt},
            x label style={yshift=3pt},
            width=\textwidth,
            height=5cm,
            legend pos=south east
            ]
            \addplot[color=scarletred1,font=\scriptsize] file {new_plot_ctm.dat};
            \addplot[color=chameleon3,font=\scriptsize] file {plot_lda.dat};
            \addplot[color=skyblue1,font=\scriptsize] file {plot_polylda.dat};
            \legend{\bf{Ours} $0.199$, LDA $0.148$, PolyLDA $0.177$}
        \end{axis}
    \end{tikzpicture}
\label{fig:compatibility_curve_abl}
\end{subfigure}\hspace*{-0.3in}
\begin{subfigure}[b]{0.3\textwidth} %
\pgfplotsset{every axis/.append style={
    font=\footnotesize,,
    line width=1pt,
    tick style={ultra thin}}}
    \begin{tikzpicture}
        \begin{axis}[
            xlabel=Recall,
            ylabel=Precision,
            ymin = 0,
            y label style={yshift=-3pt},
            x label style={yshift=3pt},
            width=\textwidth,
            height=5cm,
            legend pos=south east
            ]
            \addplot[color=scarletred1,font=\scriptsize] file {new_plot_ctm.dat};
            \addplot[color=chameleon3,font=\scriptsize] file {plot_bilstm.dat};
            \addplot[color=skyblue1,font=\scriptsize] file {plot_monomer.dat};
            \legend{\bf{Ours} $0.199$,BiLSTM $0.184$, Monomer $0.157$}
        \end{axis}
    \end{tikzpicture}
\label{fig:compatibility_curve_bs}
\end{subfigure}
\vspace*{-0.25in}
\caption{\KG{Compatibility accuracy comparisons.  Legend shows AP. Left: topic model variants, including PolyLDA styles~\cite{weilin-iccv2017}. Right: state-of-the-art compatibility models~\cite{mcauley-compatibility,bilstm}.}}
\label{fig:whole_compatibility}\vspace*{-0.2in}
\end{figure}

\myparagraph{Implementation}
We collect $3,957$ ``in the wild'' outfit images from \url{chictopia.com} to learn compatibility.  We apply the cross domain curriculum learning (\secref{attr_recognition}) to predict their attributes, then fit the topic model~\cite{scactm} (\secref{CTM}).  Given an outfit consisting of Polyvore garment product images, we predict attributes per piece then \KG{pool} them for the whole outfit, and infer the outfit's compatibility.
Since both Monomer~\cite{mcauley-compatibility} and BiLSTM~\cite{bilstm} are designed to run on \emph{per-garment catalog images}, training them on our full-body outfit dataset is not possible.  %
\myparagraph{Results}
\figref{whole_compatibility} (left) compares our compatibility to alternative topic models.  As one baseline, we extend the polylingual LDA (\textsc{PolyLDA}) style discovery~\cite{weilin-iccv2017} to compute compatibility with likelihood, analogous to our CTM-based approach.  %
 Among the three topic model variants, LDA performs worst. PolyLDA~\cite{weilin-iccv2017} learns styles across all body parts, and thus an outfit with incompatible pieces will be modeled by multiple topics, lowering its likelihood.  However, PolyLDA is still ignorant of correlation between styles, and our model fares better.

\figref{whole_compatibility} (right) compares our compatibility to Monomer~\cite{mcauley-compatibility} and BiLSTM~\cite{bilstm}.  Our model outperforms \KG{both existing techniques}. Monomer assumes pairwise relations, but many outfits consist of more than two pieces; aggregating pairwise relations fails to accurately capture an outfit's compatibility as a whole.  
Like us, BiLSTM learns compatibility from only positives, yet we perform better. This supports our idea to learn compatibility from ``in the wild'' full body images.

\figref{qual_compatibility} shows qualitative results of most and least compatible outfits scored by our model.  Its  most compatible outfits tend to include basic pieces---so-called \emph{staples}---with neutral colors and low-key patterns or materials.  Basic pieces go well with almost anything, making them strong compatibles.  Our model picks up on the fact that people's daily outfits  usually consist of at least one such basic piece.  Outfits we infer to be incompatible tend to have incongruous color palettes or mismatched fit/cut.

\subsection{Capsule wardrobe creation}
\label{sec:exp_capsule}
Having validated our compatibility module, we next evaluate our selection algorithm for capsule wardrobes.
We consider two scenarios for likely use cases: i) \emph{Adding}---given a seed outfit, optimize the pieces to add, augmenting the starter wardrobe; and ii) \emph{Personalizing}---given a set of outfits the user likes/has worn/owns, optimize a new capsule from scratch to meet his/her taste.
\myparagraph{Dataset}
To construct the pool of candidate pieces, we select \KG{$N=150$} pieces each for the \emph{outer}, \emph{upper}, and \emph{lower} layers and \KG{$N=50$} for the \emph{one piece} layer from the $7,478$ pieces in the Polyvore data.  %
The pool size represents the scale of a typical online clothing vendor.   As seed outfits, we use those that have pieces in all \emph{outer}, \emph{upper}, \emph{lower} layers, resulting in $759$ seed outfits.  \KG{We report results averaged over all $759$ seed initializations.}
We consider capsules with \KG{$T=4$} pieces in each of the $m=3$ layers.  This gives $12$ pieces and $64$ outfits per capsule.
\myparagraph{Baselines}
The baselines try to find staples or prototypical pieces, while at the same time avoiding near duplicates. 
Specifically: i) \textsc{MMR}~\cite{mmr}: widely used function in information retrieval that strikes a balance between ``relevance'' and ``diversity'', \KG{scoring items by} $\lambda Rel+(1-\lambda)Div$.  We
use our  model's $p(s | \boldsymbol{\mu},\boldsymbol{\Sigma},\beta)$ of a single piece $s$ to measure MMR relevance, 
and the visual dissimilarity between selected pieces as diversity.
ii) \textsc{Cluster Centers}: clusters the pieces in each layer to \KG{$T$} clusters and then selects a representative piece from each cluster and layer.  
We cluster with $k$-medoids in the $2048$-D feature space from the last layer of our catalog attribute CNN.

\begin{table}
\footnotesize
    \begin{center}
    \small
       \begin{tabular}{@{}l|cc@{}}                   
          & Compatibility ($\downarrow$) & Versatility ($\uparrow$)\\
          \midrule
          Cluster Center & 1.16 & 0.55 \\
          MMR-$\lambda$0.3 & 3.05 & \bf3.09 \\
          MMR-$\lambda$0.5 & 2.95 & 2.85 \\
          MMR-$\lambda$0.7 & 2.12 & 2.08 \\
          na\"\i ve greedy & 0.88 & 0.84 \\
          Iterative  & \bf0.83 & 0.78
       \end{tabular}
       \vspace*{-0.1in}
   \caption{Capsules scored by human-created gold standard.}
   \label{tab:capsule_eval} 
   \end{center}
    \vspace*{-0.3in}
\end{table}

\begin{figure}[t]
   \begin{center}
   \begin{subfigure}{\textwidth}
        \includegraphics[width=.5\linewidth]{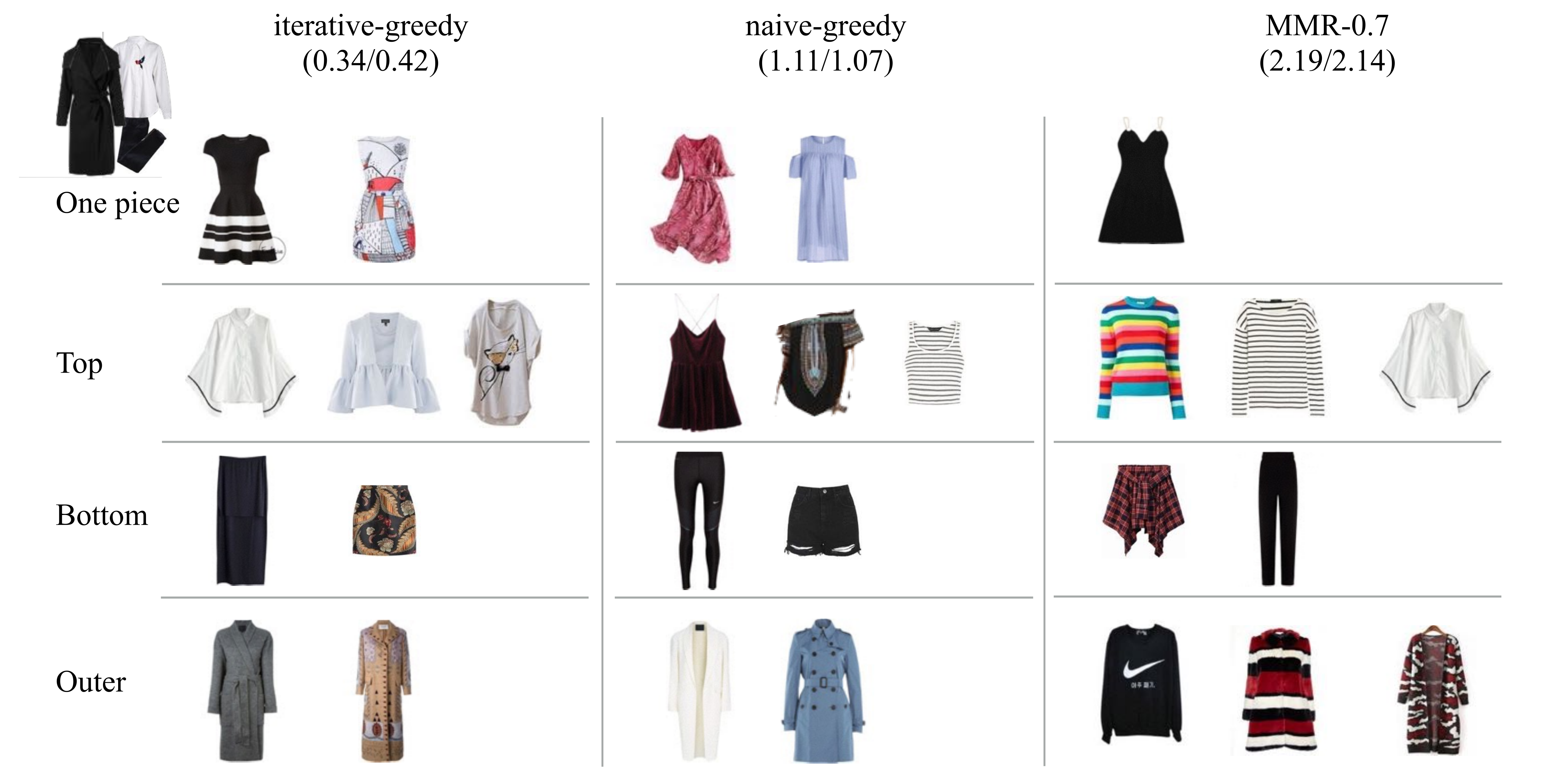}
    \end{subfigure}
    \begin{subfigure}{\textwidth}
        \includegraphics[width=.5\linewidth]{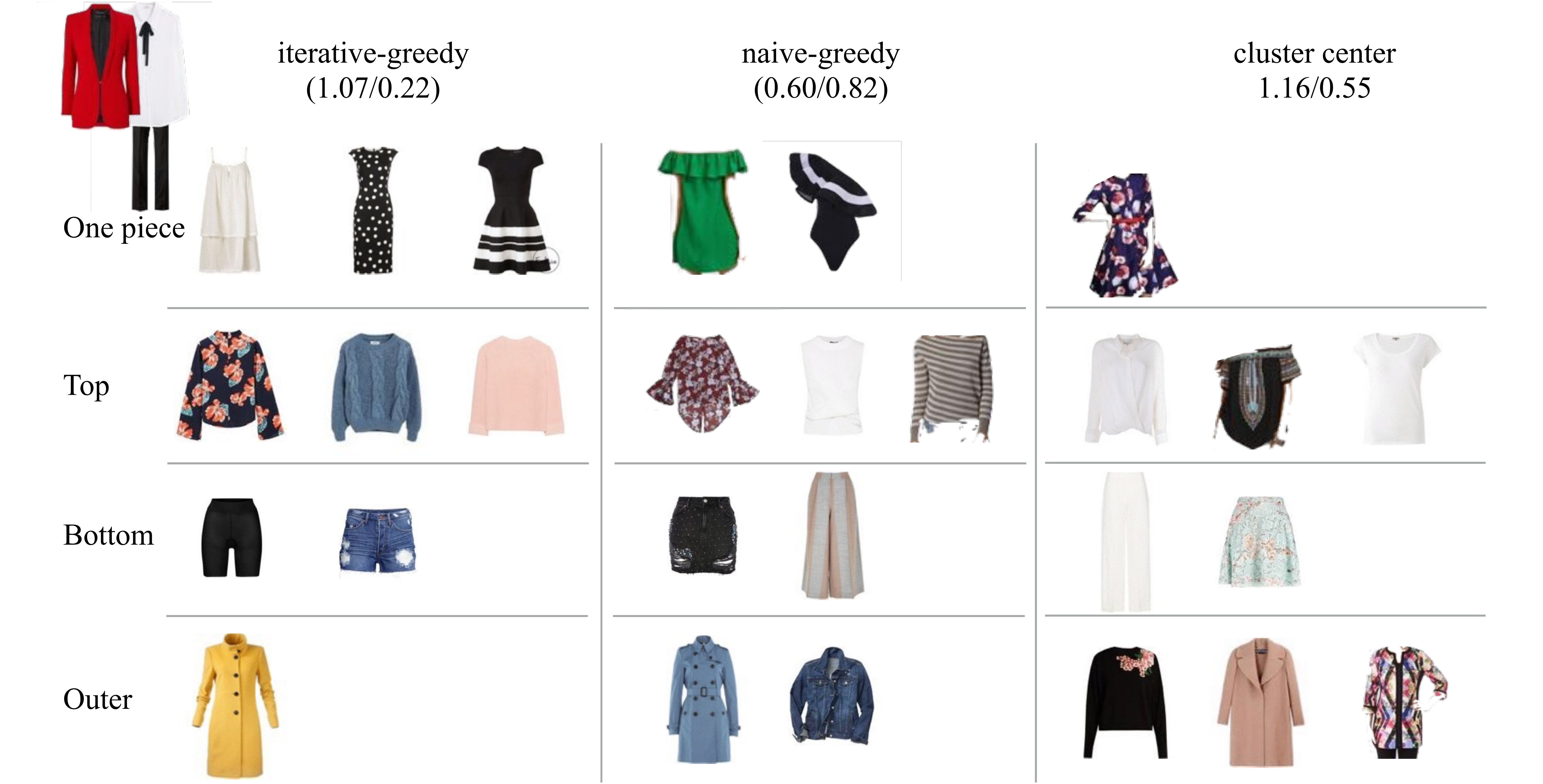}
    \end{subfigure}
    \end{center}
\vspace*{-0.2in}
   \caption{Two example capsules created by different methods for the same seed outfits (shown offset to left).     Titles show method and compatibility/versatility scores.}
   \label{fig:capsule_add}
   \vspace*{-0.2in}
\end{figure}
\myparagraph{Capsule creation quality}
First, we compare all methods by measuring how much their capsules differ from human-curated outfits.
The gold standard outfits are the Polyvore sets.  %
We measure the visual distance of each outfit to its nearest neighbor in the gold standard; the smaller the distance, the more compatible.
A capsule's compatibility is the summed distances of its outfits.
We score capsule versatility piece-wise: we compute the piece-wise visual distance per layer, and sum all distances.  %
All distances are computed on the $2048$-D CNN features \KG{and normalized by the $\sigma$ of all distances.}
We stress that both metrics are independent of our model's learned compatibility and styles; success is achieved by matching the human-curated gold standard, \emph{not} merely by optimizing our objectives $C(\mathbf{y})$ and $V(\mathbf{y})$.

\tabref{capsule_eval} shows the results.  Our iterative-greedy is nearest to the gold standard for compatibility, and MMR fares worst.  Tuning $\lambda$ in MMR as high as 0.7 to emphasize its relevance term  helps increase its accuracy, as it includes more staples.  
However, an undesirable effect of MMR (hidden by these numbers) is that for a given $\lambda$ value, the capsules are almost always the same, \emph{independent of the initial seed outfit}.   This is due to MMR's diversity term being vulnerable to outliers.  \KG{Hence, while MMR outperforms our method on versatility, it fails to create capsules coherent with the seed wardrobe (see \figref{capsule_add}, bottom right).}  %
Clustering has acceptable compatibility, but low versatility (see \figref{capsule_add}, top right).

\myparagraph{Personalized capsules}
Next we demonstrate %
our approach to tailor a capsule for a user's taste.   As a proof of concept, we select two users from \url{chictopia.com}, and use $200$ photos in their albums to learn the users' style preference (see end of Sec.~\ref{sec:CTM}). All $7,478$ pieces are treated as candidates.  \figref{capsule_personalized} shows the personalized capsules generated by our algorithm.  User 1's album suggests she prefers lady-like looks, and accordingly our algorithm creates a capsule with pastel colors and chiffon material.  In contrast, user 2 prefers street, punk looks, and our algorithm creates a capsule with denim, leather material, and dark colors. \ccc{See Supp. for a comparison to nearest neighbor image retrieval.}

\begin{table}
\small
    \begin{center}
    \begin{tabular}{@{}l|ccc}%
      & Objective & Obj./Optimal Obj.(\%)& Time \\ %
      \midrule
      Optimal & 40.8 & 100 & 131.1 \emph{sec}\\
      na\"\i ve greedy & 30.8 & 76 & 34.3 \emph{sec} \\
      Iterative & 35.5 & 87 & 57.9 \emph{sec} %
    \end{tabular}
    \vspace*{-0.1in}
   \caption{\KG{Quality vs.~run-time \KG{per capsule} for na\"\i ve and iterative greedy maximization, compared to the true optimal solution. Here we run at toy scale ($N=10$) so that optimal is tractable.  Our iterative algorithm better approximates the optimal solution, yet is much faster.  Run at a realistic scale ($N=150$), optimal is intractable ($\sim$1B hours), while our algorithm takes only 200 sec.}}%
   \label{tab:capsule_obj} 
    \end{center}
    \vspace*{-0.34in}
\end{table}

\myparagraph{Iterative submodular vs.~na\"\i ve greedy algorithm}  \KG{Next, we compare our iterative greedy algorithm---which properly accounts for the superlinear growth of capsules as garments are added---to a baseline greedy algorithm that na\"\i vely employs submodular function maximization, ignoring the combinatorics of introducing each new garment.} 
To verify that our iterative approach better approximates the optimal solution in practice, we create a toy experiment with $N=10$ candidates and $T=3$ selections per layer.  We stress the scale of this experiment is limited only by the need to compute the true optimal solution. All algorithms create capsules from scratch.
We run all methods on the same single Intel Xeon 2.66Ghz machine.

\tabref{capsule_obj} shows the results.  Our iterative algorithm achieves $87\%$ of the optimal objective function value, a clear margin better than na\"\i ve at $76\%$. On the toy dataset, solving capsule wardrobes by brute force takes $\sim2\times$ our run-time. Run at the realistic scale of our experiments above ($N=150$) the brute force solution is intractable, $\sim$1B hours per capsule, but our solution takes only 200 sec.  

\begin{figure}
   \begin{center}
   \begin{subfigure}{\textwidth}
        \includegraphics[width=.5\linewidth]{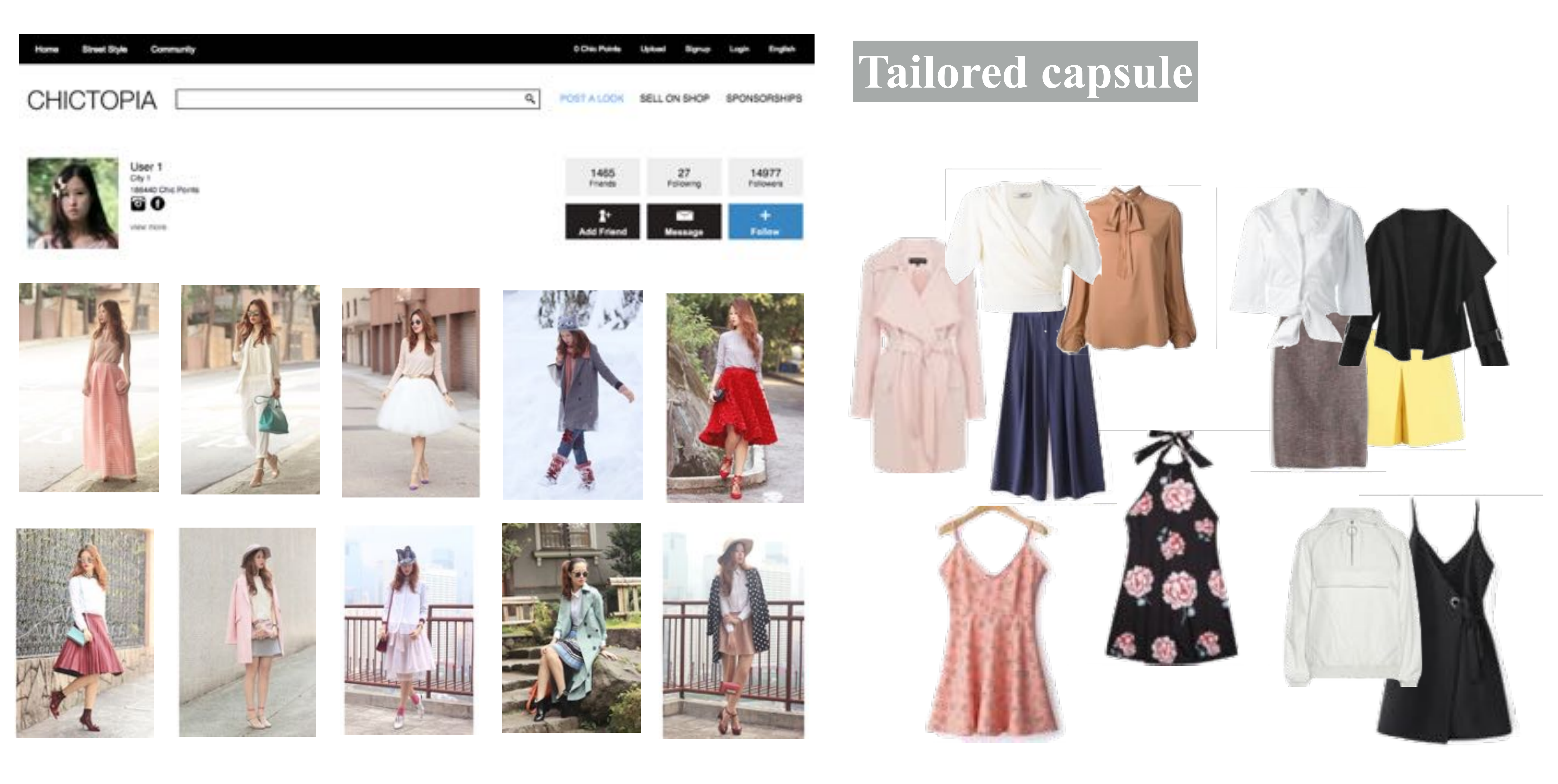}
   \end{subfigure}
    \begin{subfigure}{\textwidth}
        \includegraphics[width=.5\linewidth]{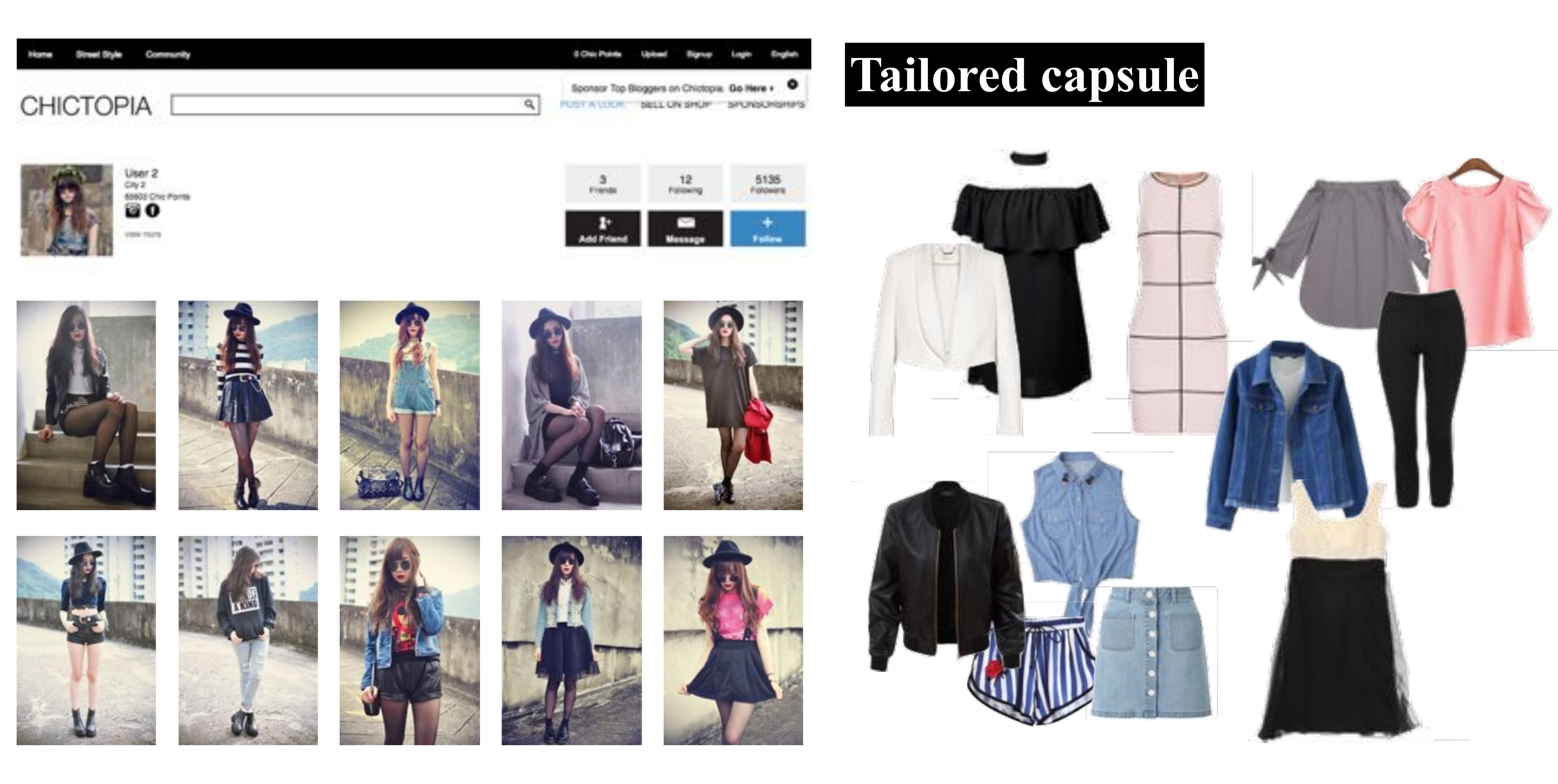}
    \end{subfigure}
    \vspace*{-0.15in}
    \caption{Personalized capsules tailored for user preference.}
   \label{fig:capsule_personalized}
   \end{center}
   \vspace*{-0.34in}
\end{figure}

\myparagraph{Comparative human subject study}
How do humans perceive the gap between iterative and na\"\i ve greedy's results?   To analyze this, we perform a human perception study with $14$ subjects.  \KG{Since the data contains women's clothing, all subjects are female, and they range in age from 20's to 60's.}  Using a Web form, we present $50$ randomly sampled pairs of capsules (iterative vs.~na\"\i ve greedy), displayed in random order per question.  %
Then for each pair, the subjects must select which is better.  See Supp.~for interface.  We use majority vote and weight by their confidence scores.    
Despite the fact that na\"\i ve greedy leverages the same compatibility and versatility functions, 59\% of the time the subjects prefer our iterative algorithm's capsules.  This shows the impact of the proposed optimization.

\vspace*{-0.1in}
\section{Conclusion}
\vspace*{-0.1in}

Computer vision can play an increasing role in fashion applications, which are, after all, inherently visual.  Our work explores capsule wardrobe generation.  The proposed approach offers new insights in terms of both efficient optimization for combinatorial mix-and-match outfit selection, as well as generative learning of visual compatibility.  Furthermore, we demonstrate that compatibility learned in the wild can successfully translate to clean product images via attributes and a simple curriculum learning method.  Future work will explore ways to optimize attribute vocabularies for capsules.

\ccc{\myparagraph{Acknowledgements:} We thank Yu-Chuan Su and Chao-Yuan Wu for helpful discussions. We also thank our human subjects: Karen, Carol, Julie, Cindy, Thaoni, Maggie, Ara, Jennifer, Ann, Yen, Chelsea, Mongchi, Sara, Tiffany. This research is supported in part by an Amazon Research Award and NSF IIS-1514118. We thank Texas Advanced Computing Center for their generous support.}

\section*{Appendix Overview}

This document consists of:
\begin{itemize}
\item [--] Proof of Claim 3.2 in Section 3.1.2 of the main paper
\item [--] Implementation details for iterative-greedy algorithm.
\item [--] Computation complexity of naive and iterative greedy algorithm in Section 3.1.2 of the main paper
\item [--] Vocabulary for predicted attributes in Section 3.3 of the main paper
\item [--] Examples of false negative generated by randomly swapping pieces in Section 4.1 of the main paper
\item [--] Qualitative example images for most and least compatible outfits scored by baseline methods in Section 4.2 of the main paper
\item [--] Qualitative example images for personalized capsules obtained by nearest-neighbor baseline.
\item [--] Interface for human subject study in Section 4.2 of the main paper
\end{itemize}

\section{Submodularity for objective function}
\begin{claim_unnumbered}{\normalfont{3.2}} When fixing all other layers (i.e., upper, lower, outer) and selecting a subset of pieces one layer at a time, the probabilistic versatility coverage function in Eqn~(3) is submodular, and the compatibility function in Eqn~(2) is modular.
\end{claim_unnumbered}
\begin{proof}
Let $A_j$ be candidate pieces from each layer $j$, where $j=\cbr{0,1,\dots,m-1}$, and $D$, $B$ be any set such that $D\subseteq B$, $D = \prod\limits_{j=0}^{m-1} A_j^D$, $B = \prod\limits_{j=0}^{m-1} A_j^B$, where $A_j^D \subseteq A_j$, $A_j^B \subseteq A_j, \forall j$. Since $D\subseteq B$, $A_j^D\subseteq A_j^B, \forall j$.

Given a layer $i$, let $s_i \in A_i\setminus A_i^B$ be the piece additionally included. Outfits introduced by including $s_i$ to $B$ and $D$ will be $O=\cbr{s_i\times\prod_{j\neq i}A_j^B}$ and $K=\cbr{s_i\times\prod_{j\neq i}A_j^D}$, respectively. Since $A_j^D\subseteq A_j^B$, $K\subseteq O$.

\renewcommand{\labelitemi}{$\blacktriangleright$}
\begin{itemize}
\item Given a layer $i$, and fixing all other layers, versatility is submodular.
{\scriptsize
\begin{align*}
v_{B\cup O}(z_i)-v_B(z_i) &= 1-\prod_{o_j \in B \cup O} (1-P(z_i|o_j))\\
&\quad- \Bigg(1-\prod_{o_j \in B } (1-P(z_i|o_j))\Bigg)\\
&= \prod_{o_j \in B } (1-P(z_i|o_j)) - \prod_{o_j \in B \cup O} (1-P(z_i|o_j))\\
&= \prod_{o_j \in B } (1-P(z_i|o_j))\Bigg(1-\prod_{o_j \in O} (1-P(z_i|o_j))\Bigg)\\
\end{align*}
}
Because $P(z_i|o_j)$ is defined as a probability, it is in the range $[0, 1]$, and therefore $(1-P(z_i|o_j)) \in
[0, 1], \forall j$. Since $D \subseteq B$, we have that $\prod_{o_j \in B } (1-P(z_i|o_j)) \leq \prod_{o_j \in D } (1-P(z_i|o_j))$. Thus,
{\scriptsize
\begin{align*}
&\prod_{o_j \in B } (1-P(z_i|o_j))\Bigg(1-\prod_{o_j \in O} (1-P(z_i|o_j))\Bigg) \\
&\leq \prod_{o_j \in D } (1-P(z_i|o_j))\Bigg(1-\prod_{o_j \in O} (1-P(z_i|o_j))\Bigg) \\
&= \prod_{o_j \in D} (1-P(z_i|o_j))\Bigg(1-\prod_{o_j \in K} (1-P(z_i|o_j))\prod_{o_j \in O \setminus K} (1-P(z_i|o_j))\Bigg)
\end{align*}
}
When $O\setminus K = \varnothing$,
{\scriptsize
\begin{align*}
&v_{B\cup O}(z_i)-v_B(z_i)\\
&\leq\prod_{o_j \in D} (1-P(z_i|o_j))\Bigg(1-\prod_{o_j \in K} (1-P(z_i|o_j))\prod_{o_j \in O \setminus K} (1-P(z_i|o_j))\Bigg)\\
&= \prod_{o_j \in D} (1-P(z_i|o_j))\Bigg(1-\prod_{o_j \in K} (1-P(z_i|o_j))\Bigg)\\
&= \prod_{o_j \in D } (1-P(z_i|o_j)) - \prod_{o_j \in D \cup K} (1-P(z_i|o_j))\\
&= v_{D\cup K}(z_i)-v_D(z_i)
\end{align*}
}
Since $K\subseteq O$, when $O\setminus K = \varnothing$, $O=K$, i.e.~$\cbr{s_i\times\prod_{j\neq i}A_j^B}=\cbr{s_i\times\prod_{j\neq i}A_j^D}$, and thus $A_j^B = A_j^D, \forall j\neq i$.
Submodularity is closed under nonnegative linear combination, and user's personalized preference for each style $i$ $w_i \geq 0$, thus $V(\mathbf{y})$ and $V'(\mathbf{y})$ are both submodular when given a layer $i$ and fixing all other layers.

\item Given a layer $i$, and fixing all other layers, compatibility is modular.\\
Since $s_i\in A_i \setminus A_i^B$, $\cbr{s_i\times\prod_{j\neq i}A_j^B} \cap\cbr{A_i^B\times\prod_{j\neq i}A_j^B} = \varnothing$, i.e.~$O\cap B = \varnothing$, and same with $D\cap K = \varnothing$.\\
By $O\cap B = \varnothing$, we get 
{\footnotesize
\begin{align*}
C(B\cup O)-C(B) &= \sum\limits_{o_j \in B\cup O} c(o_j)-\sum\limits_{o_j \in B} c(o_j)\\
&= \sum\limits_{o_j \in B} c(o_j)+\sum\limits_{o_j \in O} c(o_j)-\sum\limits_{o_j \in B} c(o_j)\\
&=\sum\limits_{o_j \in O} c(o_j)
\end{align*}
}
and $C(D\cup K)-C(K) = \sum\limits_{o_j \in K} c(o_j)$.\\
By $K\subseteq O$, we have
{\footnotesize
\begin{equation*}
\sum\limits_{o_j \in O} c(o_j) = \sum\limits_{o_j \in K} c(o_j) + \sum\limits_{o_j \in O\setminus K} c(o_j)
\end{equation*}
}
When $O\setminus K=\varnothing$,
{\footnotesize
\begin{align*}
C(B\cup O)-C(B) &= \sum\limits_{o_j \in O} c(o_j)\\
&= \sum\limits_{o_j \in K} c(o_j) + \sum\limits_{o_j \in O\setminus K} c(o_j)\\
&= \sum\limits_{o_j \in K} c(o_j) = C(D\cup K)-C(K)
\end{align*}
}
Thus $C(\mathbf{y})$ is modular when given layer $i$ and fixing all other layers.
\end{itemize}
\end{proof}

\section{Implementation details for iterative-greedy algorithm}
We set our tolerance degree $\varepsilon=0.5$, and find that our iterative-greedy algorithm typically converges after $5$ iterations. 
We use Gibbs sampling~\cite{scactm} for compatibility inference. Due to the sampling process, $p(\boldsymbol{\theta},\boldsymbol{z}|o_j,\boldsymbol{\mu},\boldsymbol{\Sigma},\beta)$ fluctuates slightly at different run times. To increase robustness, we further apply a step function on our compatibility score $c(o_j)$, so that $c(o_j)\geq \epsilon$ is mapped to $1$, and otherwise to $0$. We fix $\epsilon=-4.69$, as validated in the compatibility experiment to give the best precision-recall trade-off.

\section{Computation complexity for niave and greedy algorithms}
Both the naive greedy and iterative greedy algorithms have a computation bottleneck when computing the objective $\mathbf{obj}(\mathbf{y_{t}})$. Its complexity is decided by $N_i$ times $|\mathbf{y_{t}}|$, the size of the incrementally growing subset $\mathbf{y_{t}}$ at iteration $t$.
In the following we show the algorithms of naive and iterative greedy, and analyze their $|\mathbf{y_{t}}|$ respectively. Without loss of generality, we assume our algorithms are provided with an initial piece for each layer $i$.
\begin{algorithm}[H]
    \small
    \caption*{Naive greedy algorithm for submodular maximization, where $\mathbf{obj}(\mathbf{y}) := C(\mathbf{y}) + V(\mathbf{y})$.}
    \label{em_greedy_alg}
    \begin{algorithmic}
        \For {each time step $t=1,2,...T$}
            \State $\mathbf{y_{t-1}} = \prod\limits_{i=0}^{m-1} A_{i(t-1)}$
            \For {each layer $i=0,1,...m-1$}
                
                
                \State $s_{i}^{t} := \argmax_{s \in A_i \setminus A_{i(t-1)}} \delta_s$
                \State where $\delta_s = \mathbf{obj}(\mathbf{y_{t}}) - \mathbf{obj}(\mathbf{y_{t-1}})$
                \State where $\mathbf{y_{t}} = {\mathbf{y_{t-1}}}\cup \cbr{s\times\prod_{j \neq i} A_{j(t-1)}}$
            \EndFor
        \EndFor
    \end{algorithmic}
\end{algorithm}
We need to compute $\mathbf{obj}(\mathbf{y_{t}})$ for all $s \in A_i \setminus A_{i(t-1)}$. 
At time step $t$ and layer $i$, $|A_i \setminus A_{i(t-1)}| = N_i-(t+1)$. Since $N_i\gg (t+1)$ and is around the same scale for all $i$, we shorthand the term $N_i-(t+1)$ to $N$. Considering every candidate $s$ and the set of outfits $\cbr{s\times\prod_{j \neq i} A_{j(t-1)}}$ introduced, computation for all sets introduced by all $s$ becomes $N(t+1)^{(i-1)}$ times. Summing over all $i$ and all $t$, the total computation is $\sum\limits_{t=1}^{T}\sum\limits_{i=0}^{(m-1)}N(t+1)^{(i-1)}$ times.
{\footnotesize
\begin{align*}
\sum\limits_{t=1}^{T}\sum\limits_{i=0}^{(m-1)}N(t+1)^{(i-1)} &= N\sum\limits_{t=1}^{T}\frac{1-(t+1)^m}{1-(t+1)}\\
&=N\sum\limits_{t=1}^{T}O(t^{m-1})
\end{align*}
}
$\sum\limits_{t=1}^{T}O(t^{m-1})$ is an $(m-1)$-th power series for the first $T$ natural numbers. A closed formula at $m=4$ equals $[\frac{T(T+1)}{2}]^2$, so the final complexity will be $O(NT^4)$ for naive greedy.

\begin{algorithm}[H]
    \footnotesize
    \caption{Proposed iterative greedy algorithm for submodular maximization, \KG{where $\mathbf{obj}(\mathbf{y}) := C(\mathbf{y}) + V(\mathbf{y})$.}}
    \label{em_greedy_alg}
    \begin{algorithmic}[1]
        \State $A_{iT} := \varnothing, \forall i$
        \State $\Delta_{obj} := \varepsilon + 1$ \Comment{$\varepsilon$ is the tolerance degree for convergence}
        \State $\mathbf{obj}^{m-1}_{prev} := 0$
        \While {$\Delta_{obj}^{m-1}\geq\varepsilon$}
            \For {each layer $i=0,1,...(m-1)$}
                \State $A_{iT} = A_{i0} := \varnothing$ \Comment{Reset selected pieces in layer $i$}
                \State $\mathbf{obj}^{i}_{cur} := 0$
                \For {each time step $t=1,2,...T$}
                    \State $\mathbf{y_{t-1}} = A_{i(t-1)}\times\prod_{i' \neq i} A_{i'T}$
                    \State $s_{i}^{j_t} := \argmax_{s \in A_i \setminus A_{i(t-1)}} \delta_s$ \Comment{Max increment}
                    \State where $\delta_s = \mathbf{obj}(\mathbf{y_{t-1}}\uplus s) - \mathbf{obj}(\mathbf{y_{t-1}})$
                    \State $A_{it} := s_{i}^{j_t}\cup A_{i(t-1)}$ \Comment{Update layer $i$}
                    \State $\mathbf{obj}^{i}_{cur} := \mathbf{obj}^{i}_{cur} + \delta_{s_{i}^{j_t}}$
                \EndFor
            \EndFor
            \State $\Delta_{obj}^{m-1} := \mathbf{obj}^{m-1}_{cur}-\mathbf{obj}^{m-1}_{prev}$
            \State $\mathbf{obj}^{m-1}_{prev} := \mathbf{obj}^{m-1}_{cur}$
        \EndWhile
        \Procedure{Incremental Addition }{$\mathbf{y_t} := \mathbf{y_{t-1}}\uplus s$}
        \State $\mathbf{y_{t}^+} := s, s \in A_i \setminus A_{i(t-1)}$
        \For {$j \in \cbr{1,\dots,m}, j\neq i$}
            \If {$A_{jT}\neq\varnothing$}
                \State $\mathbf{y_{t}^+} := \mathbf{y_{t}^+}\times A_{jT}$
            \EndIf
        \EndFor
        \State $\mathbf{y_{t}} := \mathbf{y_{t-1}} \cup \mathbf{y_{t}^+}$
        \EndProcedure
    \end{algorithmic}
\end{algorithm}

Let $t_g$ denote at which iteration the while loop is. At iteration $t_g=0$, for layer $i=0$, each candidate piece $s$ will introduce a set of outfits $\cbr{s\times\prod_{j\neq i}A_{jT}}$. Since each layer has an initial piece, $|A_{jT}|=1,\forall j$, and computation for the objective value of all sets introduced by all $s$ is $N$ times. After layer $i=0$ selects $T$ pieces, $|A_{0T}| = T$, and thus layer $i=1$ computes $TN$ times objective values. After that, layer $i=2$ computes $T^2N$ times, and so on. So at $g_t = 0$, the total computation complexity is $\sum\limits_{i=0}^{m-1}T^iN$. For all iterations $t_g \geq 1$, we reset selected pieces at each layer $i$, and select $T$ pieces again, so the complexity is $T^{(m-1)}N, \forall i$. Summing over all layers, we get $\sum\limits_{i=0}^{m-1}T^{(m-1)}N = mNT^{(m-1)}$. At $m=4$, we get computation complexity per $t_g$ iteration $O(NT^3)$ for iterative greedy.

\section{Vocabulary for catalog/outfit attributes}
\tabref{predict_attr} lists the predicted attributes organized by types: \emph{pattern}, \emph{material}, \emph{shape}, \emph{collar}, \emph{article}, \emph{color}. We pair \emph{pattern}, \emph{material}, and \emph{color} with body parts to get localized attributes. Since modeling correlation between attributes (e.g.~\emph{material translucent} co-occurs with \emph{pattern lace}, \emph{neckline scoop} co-occurs with \emph{pattern graphics}) improves each individual attribute accuracy~\cite{mixmatch2015,gallagher-eccv2012}, we subsample images from each type and multilabel them for catalog attribute prediction.

\begin{table}
    \scriptsize
    \setlength{\tabcolsep}{0.4em} 
    \begin{tabular}{lllllll} 
    pattern & material & shape & collar & article & color\\
    \midrule
    crochet & translucent & skirt drape pleated & scoop & T-shirt & white \\
    camouflage & leather & skirt drape prairie & vneck & blouse & black \\
    floral & denim & skirt drape flat & square & jacket & red \\
    geo & fur & skirt length long & off-shoulder & blazer & pink \\
    horizontal striped & down & skirt length medium & sweetheart & cardigan & orange \\
    lace & & skirt length short & turtle-neck & coat & yellow \\
    leopard & & skirt shape tight & shirt collar & vest & green \\
    plaid & & skirt shape loose & & dress & blue \\
    paisley & & skirt shape full && skirt & purple \\
    plain & & pants loose && pants & brown \\
    polka dot & & pants flared && jeans & gray \\
    tribal & & pants peg-leg && leggings & beige\\
    vertical striped & & pants skinny && stocking & \\
    zebra & & pants short && boots & \\
    && ruffle shirt && shoes & \\
    && ruffle dress && sunglasses & \\
    &&&& hat & \\
    &&&& belt & \\
    &&&& scarf & \\
    &&&& bag & \\
    &&&& socks & \\
    &&&& sweater & \\
    \end{tabular}
    \caption{Predicted attributes organized by types.}
    \label{tab:predict_attr}
\end{table}

\section{Examples of false negatives}
\begin{figure}
  \begin{center}
       \includegraphics[width=\linewidth]{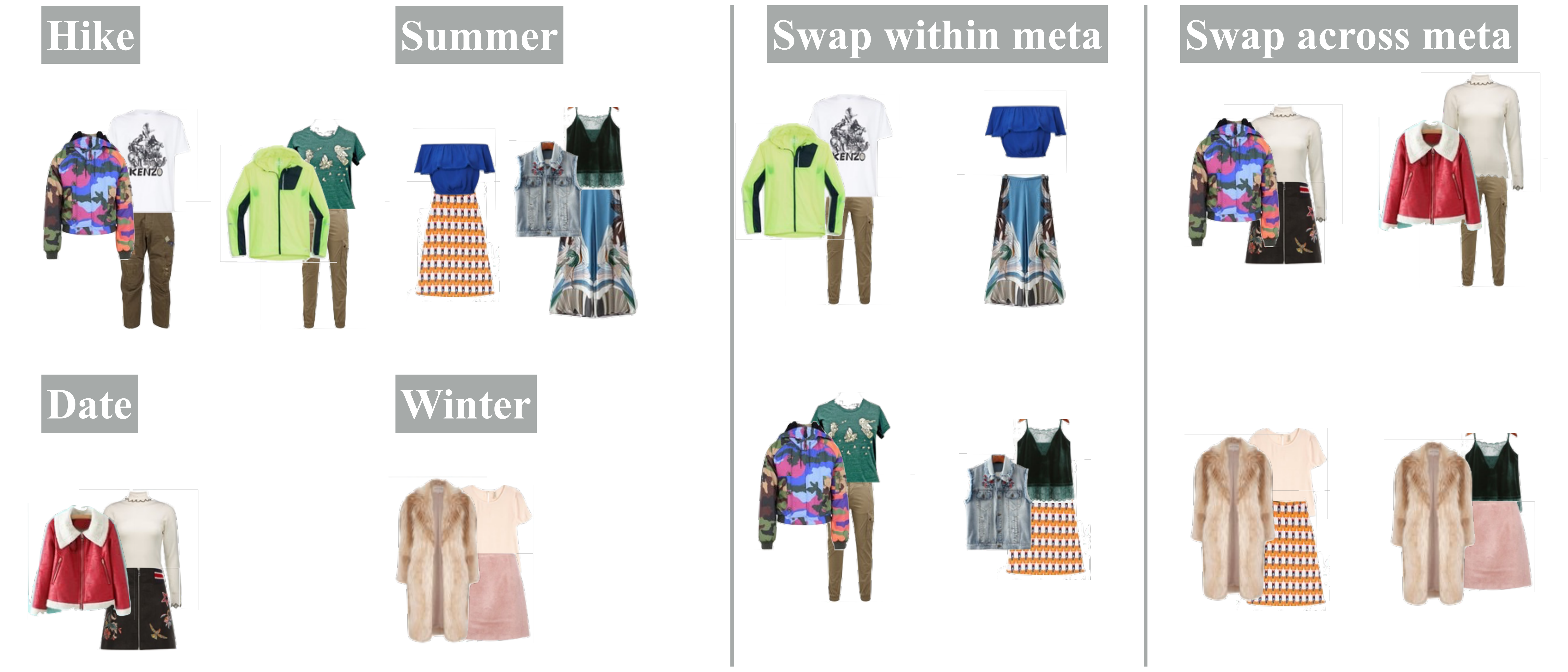}
  \end{center}\vspace*{-0.3in}
  \caption{\small{Left: outfits from exclusive meta-labels. Middle: randomly swapping pieces will form actually compatible outfits, i.e.~those swapped within the same meta-label. Right: swapping pieces across exclusive meta-labels will be closer to true negatives.}}
  \label{fig:negatives}
  \vspace{-4mm}
\end{figure}

In Section 4.1 we describe our procedure to generate negative (not compatible) outfits for evaluation.  Here we give more intuition about why this helps generate safe negatives.
In \figref{negatives} we show examples of outfits with different meta labels (\emph{season}, \emph{occasion}, \emph{function}), and show negatives generated by swapping pieces from exclusive meta-labels, comparing with negatives randomly generated.

\begin{figure*}[t!] 
  \begin{center}
       \includegraphics[width=\linewidth]{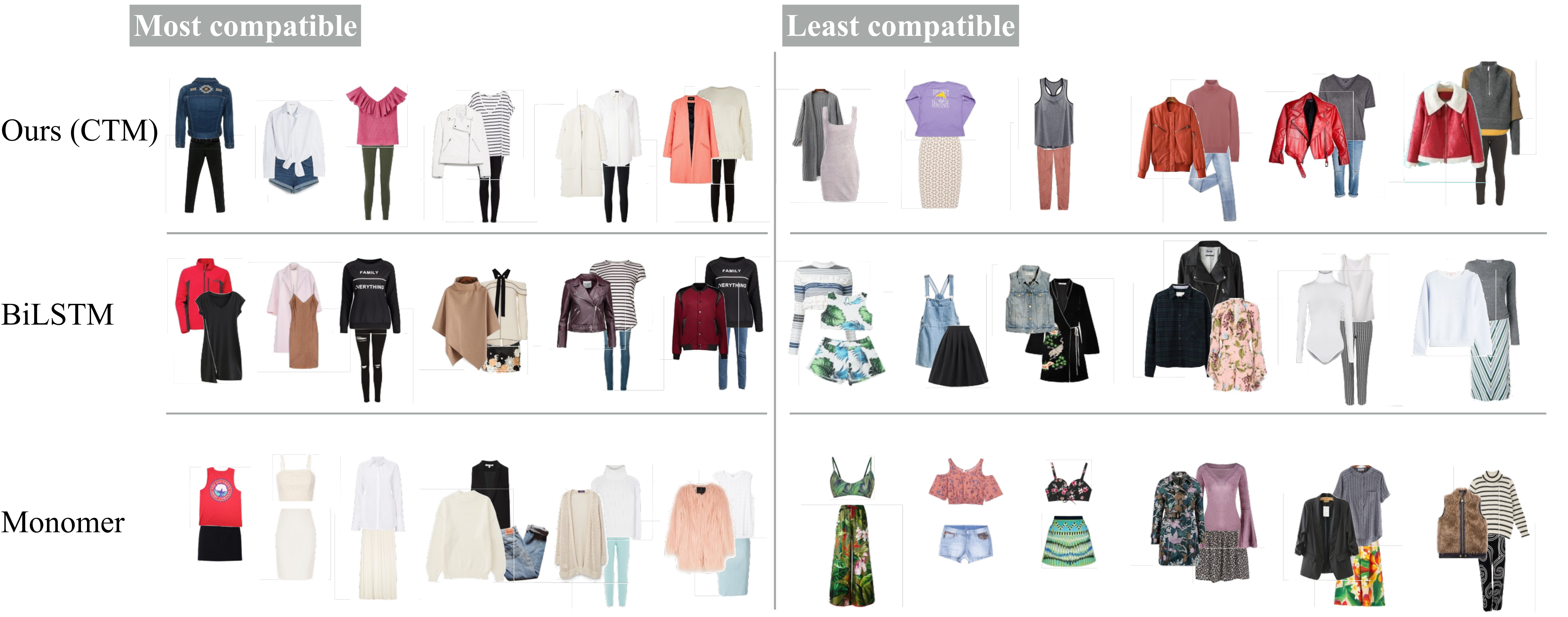}
  \end{center}\vspace*{-0.3in}
  \caption{\small{Most/least compatible outfits predicted by each method. Each row shows a method: our method tends to score outfits with staples as more compatible; BiLSTM~\cite{bilstm} scores outfits more stylish as more compatible; Monomer~\cite{mcauley-compatibility} scores outfits with white pieces as more compatible.}}
  \label{fig:compatibility}
  \vspace{-4mm}
\end{figure*}

\section{Qualitative example images for most/least compatible outfits predicted by baselines}
\figref{compatibility} shows most and least compatible outfits predicted by baselines, Monomer~\cite{mcauley-compatibility}, BiLSTM~\cite{bilstm}, along with ours (CTM). Most compatible outfits scored by us are those that consist of staples. Most compatibles scored by BiLSTM are those with a special pattern or material, which are more stylish. Most compatibles scored by Monomer contain mostly white pieces.

\section{Qualitative example images for personalized capsules obtained by nearest neighbor baseline}
We predict attributes on users' outfits and all candidate pieces, and find the visually similar pieces (measured in attribute space) to those worn on each user. The result is shown in \figref{capsule_personalized}, comparing with the result using our method. Forming capsule wardrobes by using nearest neighbor does not take compatibility nor diversity into consideration, thus the results are mainly pieces similar in cut, shape, material and color.
\begin{figure}
   \begin{subfigure}{\linewidth}
        \centering
        \includegraphics[width=.9\linewidth]{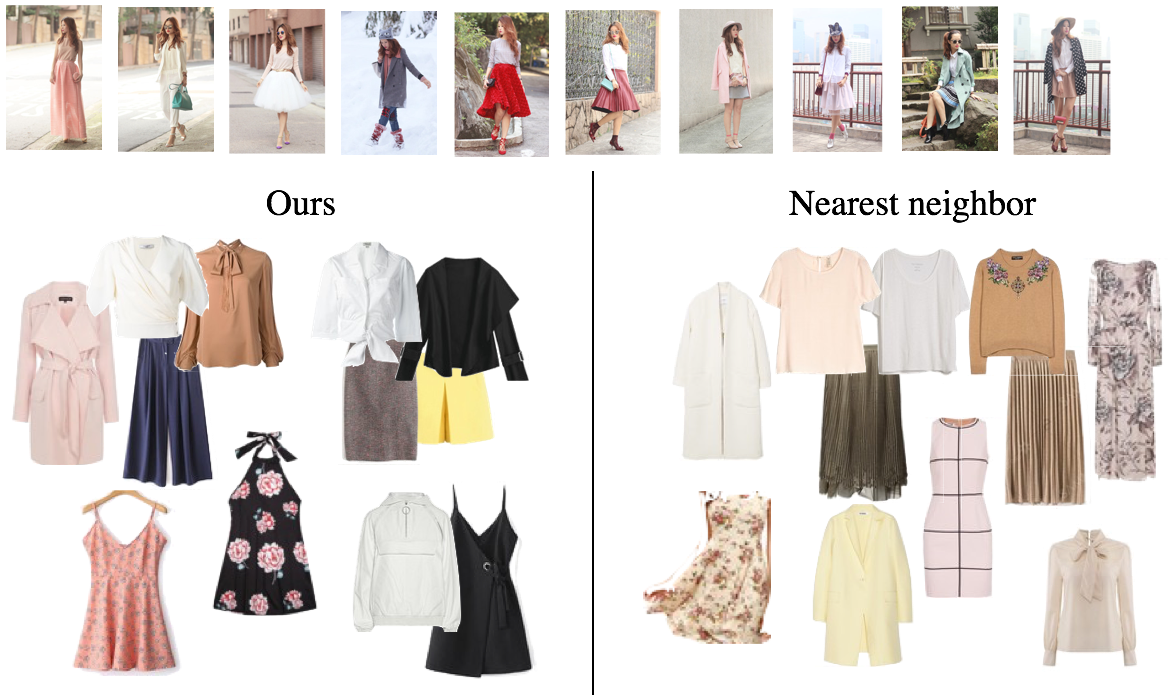}
    \vspace*{0.1in} 
   \end{subfigure}
    \begin{subfigure}{\linewidth}
        \centering
        \includegraphics[width=.9\linewidth]{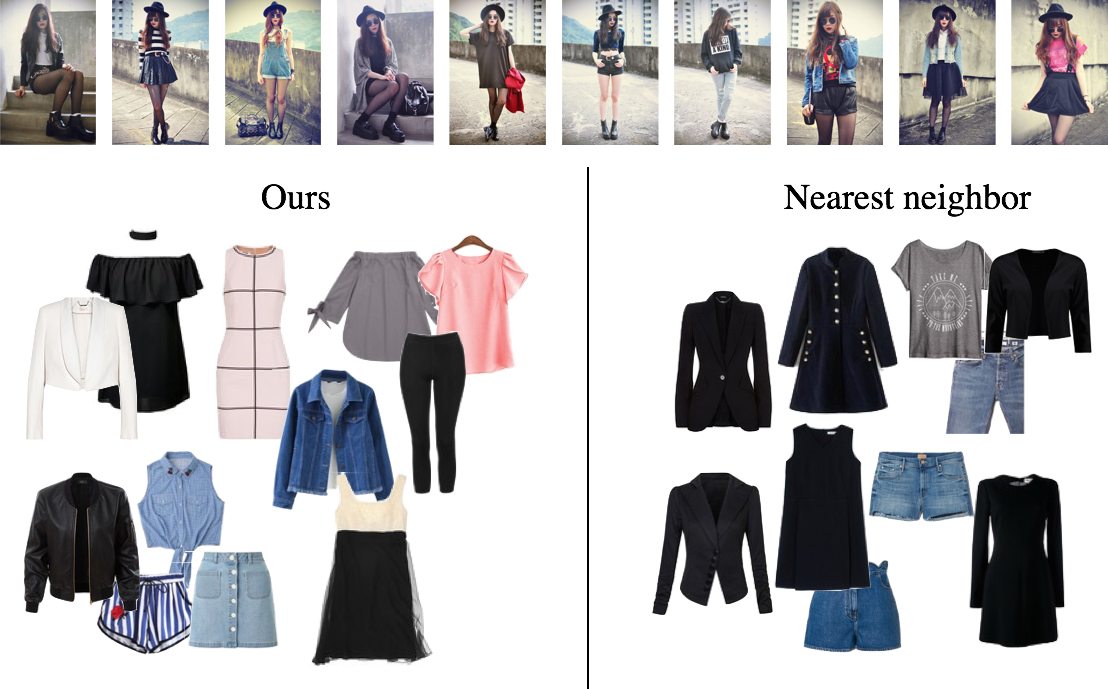}
    \end{subfigure}
    \vspace*{-0.15in}
    \caption{Personalized capsules tailored for user preference.}
   \label{fig:capsule_personalized}
   \vspace*{-0.24in}
\end{figure}

\section{Human subject interface}
\begin{figure*} 
  \vspace{-4mm}
  \begin{center}
       \includegraphics[width=.7\linewidth]{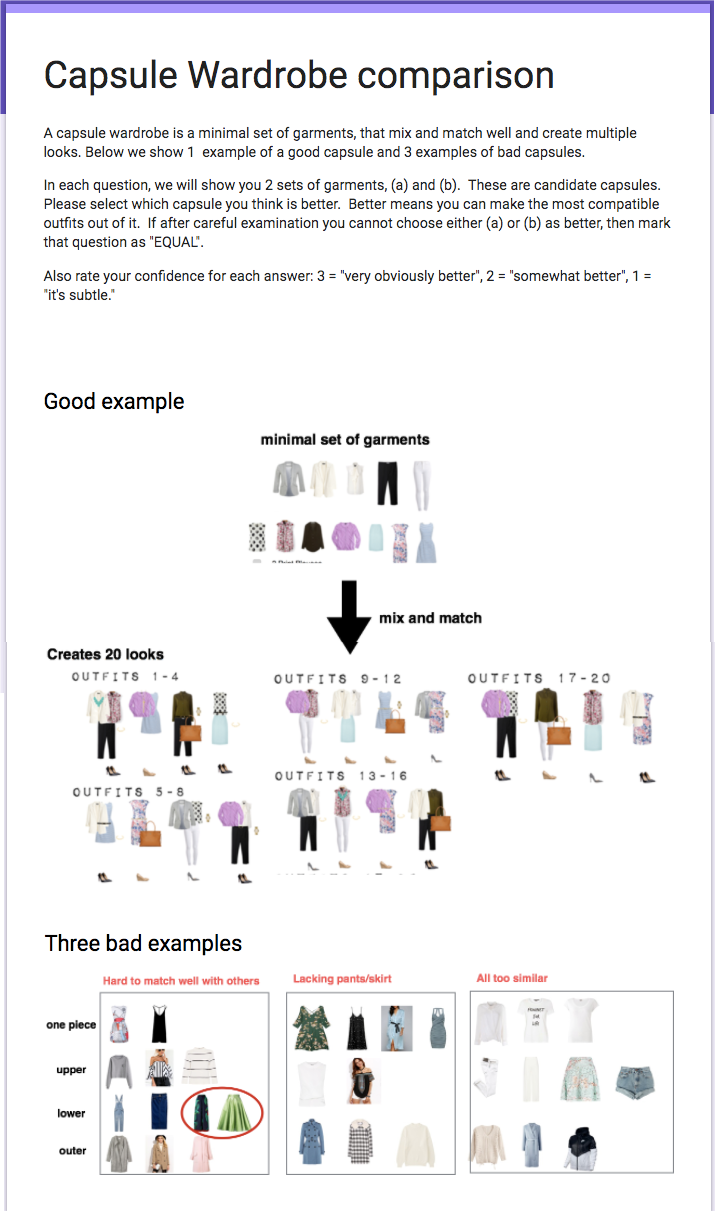}
    \caption{\footnotesize{Instructions to guide human subjects: we show textual descriptions of capsule wardrobe definitions, and visual examples of good and bad capsules.}}
    \label{fig:interface_description}
  \end{center}
\end{figure*}

\begin{figure*} 
  \vspace{-4mm}
  \begin{center}
        \includegraphics[width=.7\linewidth]{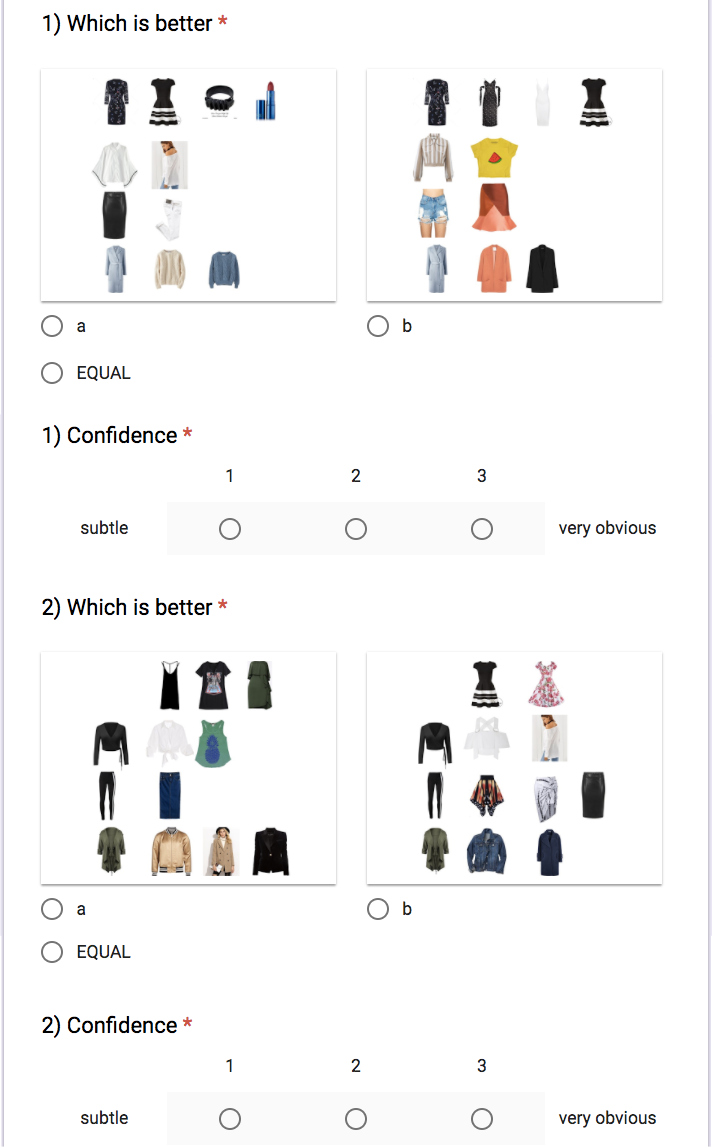}
    \caption{\footnotesize{Questions shown to subjects: (a),(b) are sampled pairs of iterative vs.~ naive greedy capsules. We encourage subjects to avoid selecting EQUAL unless the difference between two capsules is too subtle to tell. Each comparison is followed by a confidence rating for provided answer. Best viewed on pdf.}}
    \label{fig:interface_question}
  \end{center}
\end{figure*}

\figref{interface_description} and \figref{interface_question} show the interface of our human subject study on capsule wardrobes. In the instructions, we first describe the definitions of capsule wardrobes, and show examples of good and bad capsules, following explanations of why they are good and bad. In each question, we show (a) and (b) 2 candidate capsules, and ask subjects to choose which is better, where better is defined in the instructions: better capsules are those that can produce more compatible outfits. We ask subjects to avoid choosing EQUAL. Each question is also followed by confidence rating: from $1 =$ subtle to $3 = $ very obvious.

{\small
\bibliographystyle{ieee}
\bibliography{strings,egbib,refs,refs-iccv2017,refs-iccv2017b}
}

\end{document}